\documentclass{article}
\usepackage[preprint]{neurips_2026}
\usepackage{hyperref}       % hyperlinks

\usepackage{tikz}
\usetikzlibrary{arrows.meta,positioning,calc,shapes.geometric,decorations.pathreplacing,patterns}
\usepackage{pgfplots}
\usepackage{graphicx}
\usepackage{subcaption}
\usepackage{booktabs}
\usepackage{array}
\usepackage{multirow}
\usepackage{amsmath,amssymb}
\usepackage{listings}
\usepackage[table]{xcolor}
\usepackage[most]{tcolorbox}
\pgfplotsset{compat=1.18}
\usepackage{amsmath,amssymb,mathtools}
\usepackage{cleveref}
\usepackage{algorithm}
\usepackage{float}
\usepackage{algpseudocode}

\definecolor{exploitblue}{RGB}{70,130,180}
\definecolor{explorered}{RGB}{220,80,60}
\definecolor{discovergold}{RGB}{218,165,32}
\definecolor{nnmgreen}{RGB}{60,160,80}
\definecolor{lightgray}{RGB}{245,245,245}
\definecolor{codebg}{RGB}{248,248,246}
\definecolor{explcolor}{RGB}{27,120,55}      % MI / exploration additions
\definecolor{divcolor}{RGB}{33,102,172}      % Nuclear-norm additions
\definecolor{betacolor}{RGB}{179,88,6}       % β-coupling factor
\definecolor{explcolor}{RGB}{120,60,180}     % MI-advantage additions (purple, matches inspo)
\newcommand{\expl}[1]{\textcolor{explcolor}{#1}}

\lstdefinestyle{pystyle}{
    basicstyle=\ttfamily\footnotesize,
    backgroundcolor=\color{codebg},
    keywordstyle=\color{blue!70!black}\bfseries,
    commentstyle=\color{green!50!black}\itshape,
    stringstyle=\color{red!70!black},
    showstringspaces=false,
    frame=single,
    framerule=0.3pt,
    rulecolor=\color{gray!40},
    breaklines=true,
    language=Python,
    numbersep=5pt,
    xleftmargin=6pt,
    xrightmargin=6pt,
}

\newcommand{\MI}{\mathrm{MI}}
\newcommand{\Rexec}{R_{\mathrm{exec}}}
\newcommand{\Aexec}{A_{\mathrm{exec}}}
\newcommand{\geff}{\gamma_{\mathrm{eff}}}
\definecolor{methodpurple}{HTML}{7C4FC2}
\newcommand{\method}{\textsc{\textcolor{methodpurple}{UG-TTT}}}
\newcommand{\baseline}{\textsc{TTT-Discover}}

\definecolor{tblHeader}{RGB}{230,230,230}      % Slightly darker neutral header
\definecolor{tblBaseline}{RGB}{240,240,240}    % Soft grey for baseline cells
\definecolor{tblMethod}{RGB}{240,232,250}      % Soft purple tint matching methodpurple

\title{Epistemic Uncertainty for Test-Time Discovery}

\author{
Kainat Riaz$^{1*}$ \and
\textbf{Muhammad Ahmed Mohsin}$^{1*}$ \and
Ahsan Bilal$^{3}$ \and
Muhammad Umer$^{1}$ \and
Ayesha Mohsin$^{2}$ \and
Aqib Riaz$^{2}$ \and
Ali Subhan$^{2}$ \and
John M. Cioffi$^{1}$
\\[0.5em]
$^{1}$Stanford University \\
$^{2}$National University of Sciences and Technology \\
$^{3}$University of Oklahoma \\
\\
$^{*}$Equal contribution (core contributors)
}

\date{}
\makeatletter
\newcommand{\labelalias}[2]{%
\@ifundefined{r@#1}{}{%
\expandafter\let\csname r@#2\expandafter\endcsname\csname r@#1\endcsname}}
\makeatother
\begin{document}
\maketitle

%======================================================================
\begin{abstract}
%======================================================================
Automated scientific discovery using large language models relies on identifying genuinely novel solutions. Standard reinforcement learning penalises high-variance mutations, which leads the policy to prioritise familiar patterns. As a result, the maximum reward plateaus even as the average reward increases. Overcoming this limitation requires a signal that distinguishes unexplored regions from intrinsically difficult problems, which necessitates measuring disagreement across independently adapted weight hypotheses rather than relying on a single network's confidence. \method{}\footnote{\href{https://uncertainty-guided-ttt.github.io/}{Project Website} \quad|\quad \href{https://github.com/KainatRiaz98/epistemic-uncertainty-for-test-time-discovery}{Code Repository}} addresses this challenge by maintaining a small ensemble of low-rank adapters over a frozen base model. The per-token disagreement, quantified as the mutual information between ensemble predictions and weight hypotheses, isolates epistemic uncertainty and identifies positions where insufficient coverage leads to adapter divergence rather than intrinsic problem difficulty. This measure is incorporated as an exploration bonus into the policy gradient, directing the policy toward positions where persistent adapter disagreement signals low training coverage, the same frontier where genuine discovery is possible. A nuclear norm regularizer ensures the adapters remain distinct from one another, thereby preserving the exploration signal throughout
training. Across four scientific discovery benchmarks,\method{} increases the maximum reward on three tasks, maintains substantially higher solution diversity, and an ablation study confirm that the regularizer is essential for sustaining  .
\end{abstract}

\begin{figure}[!t]
\centering
\includegraphics[width=\linewidth]{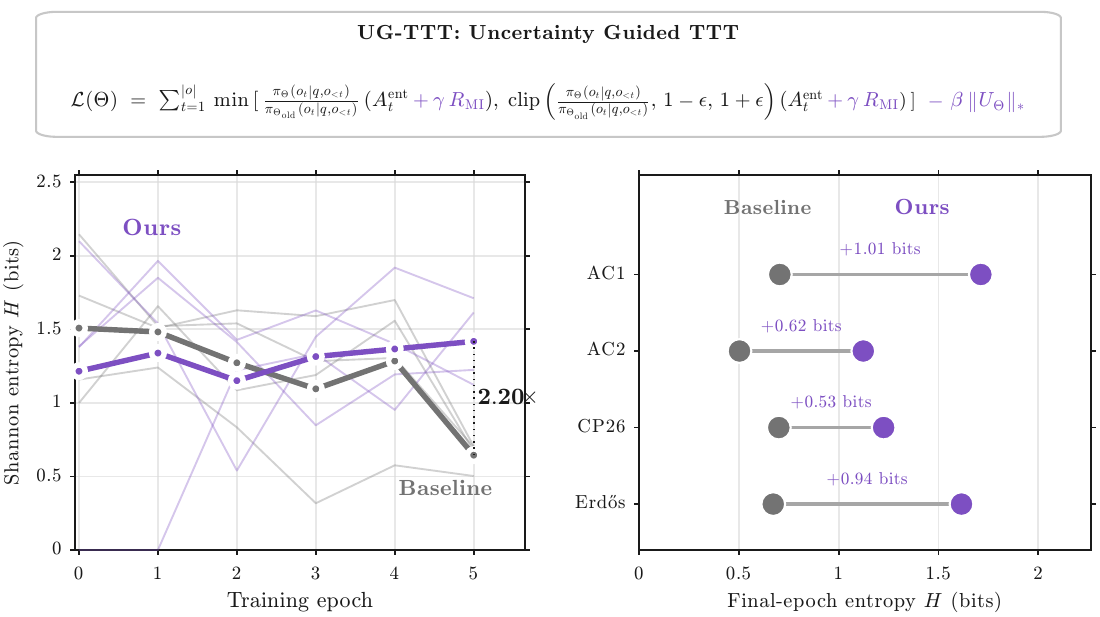}
\caption{\textbf{\method{} preserves the diversity that drives discovery.}
The training objective (top) augments the \baseline{} clipped importance-sampling (IS) surrogate
with a mutual-information (MI)-shaped advantage and a nuclear-norm regulariser (\expl{purple} terms).
\emph{Left:} Per-epoch Shannon entropy $H$ of the algorithmic family distribution,
averaged over the four benchmarks (thin lines show individual tasks); \method{} (\expl{purple})
sustains a cross-task mean $H\!\approx\!1.42$\,bits while the \baseline{} (gray)
collapses to $0.65$\,bits by epoch~5, a $2.20\times$ gap.
\emph{Right:} Final-epoch entropy gains: $+1.01$\,bits (AC1), $+0.62$ (AC2),
$+0.53$ (CP26), $+0.94$ (Erd\H{o}s), directly enabling the $R_{\max}$ improvements
of Sec.~\ref{sec:experiments}.}
\label{fig:pageone}
\end{figure}

%======================================================================
\section{Introduction}
\label{sec:intro}
%======================================================================

Large language models (LLMs) are increasingly deployed not merely as assistants but as active participants in scientific discovery, proposing hypotheses, generating candidate algorithms, and searching vast combinatorial spaces that were previously accessible only through expert intuition \cite{Qi2023Large, Yuksekgonul2026Learning, Rankovic2025Large}. Systems such as AlphaEvolve~\cite{alphaevolve} and \baseline{} have shown that an LLM coupled with a verification loop can rediscover and occasionally improve on human-designed solutions for open mathematical problems. Nevertheless, these systems consistently exhibit an inherent upper bound: while average performance can be enhanced by optimising the expected reward, the maximum attainable reward, which is critical for novel discovery, tends to plateau prematurely. This phenomenon is a direct consequence of mode collapse, where the model converges on simple, 'safe' solutions found in the training data, and a paucity of dense intrinsic signals to guide the search through the sparse rewards of the knowledge frontier\cite{Yuksekgonul2026Learning, Thomas2025TestTime}. Understanding why this plateau is structurally inherent to the expected-value objective, rather than an incidental failure of search, is the starting point of our work.

\textbf{Discovery is an out-of-distribution task.} A genuine discovery is a solution that lies beyond prior human knowledge and, consequently, outside any pretrained model’s training distribution. This framing has an immediate consequence: any method that searches high-probability regions of a model’s learned distribution primarily performs retrieval or exploitation. However, scientific discovery demands ideas beyond the training data. Therefore, progress hinges on a key question: how can we push an LLM to navigate its knowledge boundaries, where its initial predictive confidence is low, but the potential for meaningful epistemic shift is high?

\textbf{Reinforcement learning as a reward.} Existing test-time training frameworks use Predictor + Upper Confidence Bound applied to Trees (PUCT)~\cite{rosin2011puct} to select parent solutions, after which the policy generates mutations for scoring by a verifier. The failure to improve the maximum reward is structural, originating in the update rule itself: the standard reinforcement learning objective maximizes expected reward, which rewards predictive reliability and penalizes variance \cite{Cheng2025Reasoning, Li2025Know}. At the gradient level, this installs a mechanistic bias toward mutations with well-estimated outcomes, filtering out uncertain but potentially high-payoff directions. The consequence is diversity collapse, where the model converges on a narrow set of low-risk mutations, accumulating exploitative gains in average reward at the expense of the singular breakthrough state that discovery actually requires \cite{Yuksekgonul2026Learning, Thomas2025TestTime}.

\textbf{Why disagreement is the right signal.} Existing fixes for diversity collapse, including entropy bonuses, confidence-based difficulty weighting, and uncertainty-guided reasoning budgets \cite{Cheng2025Reasoning, Li2025Know}, all derive their signal from a single trained network, which cannot separate \emph{aleatoric uncertainty} (irreducible input ambiguity) from \emph{epistemic uncertainty} (gaps in what the model has learned).This is because a single network produces high entropy for both signals; distinguishing them requires measuring weight-space disagreement, a signal absent in single trajectories \cite{Hullermeier2019Aleatoric, Balabanov2024Uncertainty, AbbasiYadkori2024To}. For discovery, this conflation is fatal: the policy spends compute on ambiguous-but-familiar territory rather than the genuine knowledge frontier where breakthroughs live. AutoDiscovery~\citep{autodiscovery2025} compounds the problem further, deriving Bayesian surprise from a single frozen LLM and therefore accumulating no domain knowledge as discoveries proceed. Bayesian theory resolves the confound: mutual information between predictions and model parameters isolates the epistemic component as disagreement among independently adapted weight hypotheses \cite{Shorinwa2024A, AbbasiYadkori2024To}, so ensemble members that agree indicate mere ambiguity, while members that disagree mark a region the model has genuinely not learned. LoRA ensembles realize this decomposition without the cost of full model replication, treating each low-rank adapter as an independent posterior sample over a shared frozen base \cite{Wang2023LoRA, Balabanov2024Uncertainty}. The remaining obstacle is collapse: jointly trained adapters share data and objective and thus converge toward identical weight configurations, driving mutual information to numerical zero \cite{zhai2023uncertaintypenalizedreinforcementlearninghuman}, a degeneracy our measurements confirm within two epochs.

\textbf{Contributions.} We introduce Uncertainty-Guided Test-Time Training (\method{}), an effective framework for autonomous scientific discovery that overcomes the structural limitations of existing test-time systems. Our method extends the test-time discovery objective of the \baseline{} baseline~\cite{Yuksekgonul2026Learning} with three coupled contributions: (i) An ensemble of $K$ LoRA adapters over a shared frozen base, whose per-token disagreement, quantified as mutual information $\mathrm{MI}_t = H(\bar{p}(\cdot\mid o_{<t})) - \tfrac{1}{K}\sum_{k=1}^{K} H(p_k(\cdot\mid o_{<t}))$ where $H$ is the Shannon entropy, $p_k(\cdot\mid o_{<t})$ is the predictive distribution of the $k$-th adapter given previous tokens $o_{<t}$, and $\bar{p}_t = \tfrac{1}{K}\sum p_k$ is the arithmetic mixture distribution, supplies a Bayesian epistemic-uncertainty signal that localises positions the model has not yet resolved. (ii) A shaped advantage that injects this signal into the policy gradient with a gain coupled to the entropic temperature $\beta$, so exploration pressure rises automatically as the policy sharpens. (iii) A nuclear-norm regulariser on the stacked adapter down-projections, chosen because its global maximum is provably attained at configurations in which each adapter occupies a mutually orthogonal input subspace (Prop.~2). This is the precise condition that keeps the BALD decomposition non-trivial, and is a property that prior LoRA regularisers, focused on redundancy reduction, do not provide. Empirically \method{} sustains 3 to 4 orders of magnitude higher ensemble mutual information than an unregularised baseline, retains $1.1$ to $1.7$ bits of solution-family entropy where the baseline converges to a single family, and improves the maximum reward on three of four discovery benchmarks from the \baseline{} baseline at strictly lower token cost (Fig.~\ref{fig:pageone}).

\section{Uncertainty-guided test-time training}
\label{sec:method_full}
\label{sec:method}

A discovery-oriented policy must direct its sampling budget toward
positions the model has not learned to score, not toward positions
that are intrinsically ambiguous; the predictive entropy of a single
network does not separate these two regimes
\citep{Hullermeier2019Aleatoric, AbbasiYadkori2024To}. The Bayesian
Active Learning by Disagreement (BALD) identity isolates the
epistemic component as the mutual information between the predictive
distribution and the weight posterior, which vanishes when ensemble
members agree and grows when they place mass on incompatible
continuations \citep{bald, cao2021bayesianactivelearningdisagreements}.
We estimate this quantity from a small ensemble of low-rank
adapters, use it as a per-token exploration term in the policy
gradient, and constrain the adapter parameters to keep the
estimator non-degenerate throughout training.

\subsection{Setup}
\label{sec:method_overview}

An instance of LLM-driven discovery is an environment
$\mathcal{E}=(d,\mathcal{S},\mathcal{A},R,\mathcal{T})$ with
natural-language task description $d$, solution space $\mathcal{S}$,
action set $\mathcal{A}$ of next-token continuations, deterministic
verifiable reward
$R:\mathcal{S}\!\times\!\mathcal{A}\!\to\!\mathbb{R}$ returned by a
program checker, and transition $\mathcal{T}$ that appends an action
$a$ to a parent state $s$. A policy $\pi_\theta(a\mid q,s)$
generates a response $o=(o_1,\ldots,o_{|o|})$ of thinking tokens
and executable code, conditioned on a prompt $q$ embedding
$(d,s)$. The metric of interest is the per-instance maximum reward
$\max_{(s,a)} R(s,a)$: a single breakthrough dominates, and
improving the average without moving the maximum has not advanced
discovery.

At each training step we sample $G$ rollouts per parent state,
score each with $\Rexec^{(i)}$, and form leave-one-out advantages
$A_i$ through an adaptive temperature $\beta(s)$ computed from
$\Rexec$ alone (parent-state selection, the KL-budgeted bisection
for $\beta(s)$, and the clipped importance-sampling loss
$\mathcal{L}_{\mathrm{PG}}$ are in Appendix~\ref{app:baseline}).
The temperature $\beta(s)$ grows monotonically as the within-group
reward distribution sharpens and therefore acts as a
state-dependent measure of policy confidence, a property
Sec.~\ref{sec:method_mi_shaping} uses to couple the exploration
coefficient to the onset of mode contraction.

\subsection{Token-level epistemic uncertainty from a LoRA ensemble}
\label{sec:method_ensemble}
\label{sec:method_mi}

We replace the single \baseline{} adapter with $K$ low-rank
perturbations sharing a frozen base,
\begin{equation}
    \Theta^{(k)} \;=\; \Theta_{\mathrm{base}} + B^{(k)} A^{(k)},
    \qquad A^{(k)}\!\in\!\mathbb{R}^{r\times d_{\mathrm{in}}},\ B^{(k)}\!\in\!\mathbb{R}^{d_{\mathrm{out}}\times r},\ r\!\ll\!\min(d_{\mathrm{in}},d_{\mathrm{out}}),
    \label{eq:lora}
\end{equation}
for $k\in\{1,\ldots,K\}$. The $K$ adapters are read as samples from a
variational posterior $q(\theta\mid\mathcal{D})$ over the shared base,
and at every position $t$ the corresponding posterior predictive is the
arithmetic mixture
$\bar p_t = \tfrac{1}{K}\sum_k p_{\theta_k}(\cdot\mid q,o_{<t})$. With
this construction the standard Bayesian Active Learning by Disagreement
(BALD) decomposition isolates the epistemic component of the predictive
uncertainty at $t$:
\begin{equation}
    \MI_t \;=\; H\!\left(\bar p_t\right) \;-\; \frac{1}{K}\sum_{k=1}^{K} H\!\left(p_{\theta_k}(\cdot\mid q,o_{<t})\right),
    \label{eq:true_mi}
\end{equation}
\labelalias{eq:true_mi}{eq:mi_token}%
where $H$ is Shannon entropy over the
vocabulary~\citep{bald,cao2021bayesianactivelearningdisagreements,uncertainty_survey}.
$\MI_t$ is non-negative, vanishes exactly when every member assigns
identical mass to the next token, and grows precisely when members place
mass on incompatible alternatives, the operational hallmark of a
position the model has not yet resolved. Subtracting the mean per-member
entropy strips out the in-member (aleatoric) component, leaving only
disagreement attributable to the spread of the posterior over weights.
Both terms are evaluated on the full $(K,\mathrm{chunk},V)$ logit tensor
inside a chunked LM head; a memory-bounded implementation is described
in App.~\ref{app:implementation}.

The policy-gradient update operates on whole rollouts, but most positions
in a code rollout are syntactic boilerplate, such as indentation, brackets and the
keywords, on which the adapters trivially agree. We therefore reduce
$\MI_t^{(i)}$ to a single scalar per rollout by averaging only over the
most uncertain positions,
\begin{equation}
    U_i \;=\; \operatorname*{top\text{-}7\%\text{-}mean}_{t \in [1,|o_i|]} \MI_t^{(i)},
    \label{eq:top7}
\end{equation}
which makes $U_i$ length-robust and concentrates it on the small
minority of decoding steps that carries the meaningful signal. \citet{adadec} report that their entropy-triggered pause-and-rerank mechanism fires on $3.95\%$--$11.88\%$ of decoding steps across eight LLMs spanning $0.6$B--$8$B parameters, identifying these as the high-uncertainty positions where the ground-truth token is most often mis-ranked. We adopt $7\%$ as a heuristic order-of-magnitude choice within this range; full calibration is in App.~\ref{app:implementation}.

Two sampling decisions complete the picture. Rollouts are generated
\emph{round-robin}, so each adapter produces $G/K$ rollouts per group
and the population of generators stays balanced across training.
Scoring then evaluates \emph{every} rollout through \emph{all} $K$
adapters in a single chunked forward pass, so $\MI_t$ is well-defined
for every token of every rollout regardless of which adapter generated
it; each adapter therefore receives gradient signal from all $G$
rollouts (full schedule in Alg.~\ref{alg:ttt_mi}, App.~\ref{app:algorithm}).

\subsection{Uncertainty-shaped advantage}
\label{sec:method_mi_shaping}

Two transformations are required before $U_i$ can serve as an
additive correction to $A_i$. First, the raw scale of $U_i$ varies
by orders of magnitude across tasks and across training epochs, so
a fixed coupling coefficient would over-explore on one benchmark
while under-exploring on another; we therefore standardise $U_i$
within each rollout group. Second, a constant exploration weight maintains uniform pressure
throughout training, whereas mode collapse occurs precisely in the
regime where $\beta(s)$ grows large; we therefore couple the
exploration coefficient to $\beta(s)$, so that exploration pressure
scales monotonically with policy sharpness.

We address both with a shaped advantage. Within each group we centre
and clip $U$,
\begin{equation}
    \tilde U_i \;=\; \mathrm{clip}\!\left(\frac{U_i - \overline{U}}{\sigma_U},\; -3,\; 3\right),
    \qquad \overline{U}=\tfrac{1}{G}\sum_i U_i,\ \ \sigma_U = \sqrt{\tfrac{1}{G-1}\textstyle\sum_i(U_i-\overline U)^2},
    \label{eq:mi_standardise}
\end{equation}
with the convention that $\tilde U_i\equiv 0$ whenever $\sigma_U=0$ (the degenerate case in which all rollouts in the group share an identical $U_i$). This produces a zero-mean unit-variance signal of common scale across
tasks. The threshold $\pm 3$ follows the standard three-sigma
convention for robust advantage normalisation and lies just above the
bound $|z_i|\le\sqrt{G-1}$ that Cauchy--Schwarz imposes on the
centred unbiased z-scores, so for $G\le 10$ the clip is provably
inactive (Proposition~1) and acts as a forward-compatible safeguard
for $G\ge 11$, where $\sqrt{G-1}>3$. We then couple the exploration coefficient to
the entropic temperature,
\begin{equation}
    \geff(s) \;=\; \alpha \cdot \min\!\left(\frac{\beta(s)}{\beta_{\mathrm{ref}}},\; \gamma_{\max}\right),
    \label{eq:gamma_eff}
\end{equation}
where $\alpha$ is a base rate, $\beta_{\mathrm{ref}}$ is a reference
temperature, and $\gamma_{\max}$ is a safety clip. The shaped advantage
that replaces $A_i$ in the per-adapter policy-gradient loss
$\mathcal{L}_{\mathrm{PG}}^{(k)}(\theta_k)$ is
\begin{equation}
    A_i^{\mathrm{shaped}} \;=\; A_i \;+\; \geff(s)\cdot\overline{|A|}\cdot\tilde U_i\,,
    \qquad \overline{|A|}\;=\;\frac{1}{G}\sum_j |A_j|,
    \label{eq:a_shaped}
\end{equation}
where the prefactor $\overline{|A|}$ rescales the exploration bonus to
the same magnitude as the within-group reward signal. A near-uniform
group with small $\overline{|A|}$ therefore damps the exploration
bonus proportionally, while a decisive group permits it to compete
with the executable reward advantage $\Aexec$.

\textbf{Remark 1}~(\emph{$\beta$-coupling aligns exploration with collapse onset}).
$\beta(s)$ increases monotonically as the within-group rewards equalise, and $\beta\!\to\!\infty$ marks the limit in
which the policy contracts onto a single template.
Equation~\eqref{eq:gamma_eff} scales $\geff$ at the same rate, so
exploration pressure intensifies with policy sharpening rather than
along a fixed schedule, and the bonus remains attenuated while the
policy retains mass across modes.

\textbf{Proposition 1}~(\emph{Bounded scale perturbation of the shaped advantage}).
\emph{Let $z_i=(U_i-\overline U)/\sigma_U$ and $\mathcal{C}=\{i:|z_i|>3\}$. Then $\bigl|\sum_i A_i^{\mathrm{shaped}} - \sum_i A_i\bigr| \le \geff(s)\,\overline{|A|}\,|\mathcal{C}|(\sqrt{G-1}-3)$, and the bound vanishes whenever the clip is inactive ($\mathcal{C}=\varnothing$).}
The bonus is intentionally biased toward high-MI rollouts; this
directional bias \emph{is} the exploration mechanism. Proposition~1
(proof in App.~\ref{app:proofs}) bounds only the residual scale
perturbation, which is empty in practice: $U_i$ rarely concentrates
beyond three group-standard-deviations from the mean for the group
sizes we use ($G\!\le\!64$).

\subsection{Subspace-diversity regulariser}
\label{sec:method_nnm}

The objective so far does not preclude the $K$ adapters from
converging to a common solution. Each adapter receives identical
rollouts and shaped advantages, so identical gradients drive their
parameters toward a shared optimum and any initial disagreement
erodes. In
pilot runs the row spaces of the down-projections aligned within two to
three epochs (pairwise cosine $\to 1$), $U_i$ fell below $10^{-3}$
(Table~\ref{tab:ablation}), and the entire MI bonus quietly collapsed to
zero. The exploration mechanism of the previous subsection then
degraded silently into ordinary RL on $K$ tied adapters.

To prevent this we maximise the nuclear norm of the per-layer stacked
down-projections (Nuclear-Norm Maximisation, NNM). For each tracked
linear layer $\ell$, write
\begin{equation}
    W_\ell \;=\; \bigl[A^{(1)}_\ell;\, A^{(2)}_\ell;\, \ldots;\, A^{(K)}_\ell\bigr] \;\in\; \mathbb{R}^{Kr\times d_{\mathrm{in}}},
    \qquad
    \mathcal{L}_{\mathrm{NNM}} \;=\; -\,\frac{1}{L}\sum_{\ell=1}^{L}\,\bigl\|W_\ell\bigr\|_{*},
    \label{eq:Wstack}
\end{equation}
\labelalias{eq:Wstack}{eq:nnm_loss}%
where $A^{(k)}_\ell$ is the down-projection of adapter $k$ at layer
$\ell$, $\|\cdot\|_*$ is the nuclear norm (sum of singular values),
and $L$ is the number of tracked layers; the negative sign converts
minimisation into a maximisation of $\sum_\ell \|W_\ell\|_*$. We apply
the regulariser to the down-projections $A^{(k)}$ only and leave the
up-projections $B^{(k)}$ unconstrained, because $A^{(k)}$ controls
\emph{which} input directions an adapter attends to while $B^{(k)}$
must remain free to deliver the output magnitude the RL signal demands;
joint regularisation oscillates in pilot runs and in~\citet{uprlhf}
(App.~\ref{app:implementation}).

What makes this the right choice rather than any other low-rank
penalty is geometric:

\textbf{Proposition 2}~(\emph{The global maximum is subspace-orthogonal}).
\emph{Fix a common adapter-wise Frobenius norm
$\|A_\ell^{(k)}\|_F = c$ for all $k\in\{1,\ldots,K\}$. Whenever
$Kr\le d_{\mathrm{in}}$, the global maximum of $\|W_\ell\|_*$ on this
constraint set consists of $K$ adapter blocks
$\{A_\ell^{(k)}\}_{k=1}^K$ whose row spaces are mutually orthogonal in
$\mathbb{R}^{d_{\mathrm{in}}}$, and each block has $r$ equal singular
values, all equal to $c/\sqrt{r}$.}

A proof appears in App.~\ref{app:proofs}. In the equal-norm regime that AdamW with shared LoRA initialisation approximately maintains, 
the approximate maximisers are configurations where each adapter attends to a distinct rank-$r$ input subspace. This is the geometric
condition under which the BALD decomposition in Eq.~\eqref{eq:true_mi}
remains non-trivial: members trained on overlapping input subspaces
drift back to identical predictive distributions and zero out $\MI_t$,
which is exactly the failure mode we documented above. Existing LoRA
regularisers, including those targeting redundancy reduction or
low-rank stability, do not provide this guarantee because their stated
objectives concern individual matrices rather than the joint geometry
of the ensemble.

\subsection{Training objective}
\label{sec:method_loss}
The per-adapter policy-gradient loss with shaped advantage, a
frozen-base KL anchor, and the diversity regulariser of
Eq.~\eqref{eq:Wstack} combine into a single per-step objective,
\begin{equation}
    \mathcal{L}_{\mathrm{total}} \;=\;
    \underbrace{\frac{1}{K}\sum_{k=1}^{K}\mathcal{L}_{\mathrm{PG}}^{(k)}(\theta_k)}_{\text{ensemble RL}}
    \;+\;
    \underbrace{\lambda_{\mathrm{KL}}\,\mathrm{KL}\!\bigl(\pi_\theta\,\|\,\pi_{\mathrm{base}}\bigr)}_{\text{base anchor}}
    \;+\;
    \underbrace{\lambda_{\mathrm{NNM}}\,\mathcal{L}_{\mathrm{NNM}}}_{\text{diversity}}.
    \label{eq:total_loss}
\end{equation}
Algorithm~\ref{alg:ttt_mi} iterates over the $K$ adapters
sequentially using the parallel multi-LoRA training framework of~\citet{mlora} to bound activation memory (App.~\ref{app:algorithm}),
and the per-token reward-shaping estimator used for the KL anchor
follows~\citet{tttdiscover} (App.~\ref{app:kl_adjust}).

\textbf{Proposition 3}~(\emph{\baseline{} as a special case}).
\emph{If either (a) $K\!=\!1$, or (b) all $K$ adapters share identical
weights at every step, and additionally $\lambda_{\mathrm{NNM}}\!=\!0$,
then $\mathcal{L}_{\mathrm{total}}$ in Eq.~\eqref{eq:total_loss}
reduces to the policy-gradient objective of~\citet{Yuksekgonul2026Learning}.}
A proof appears in App.~\ref{app:proofs}.

%======================================================================
\section{Experiments}
\label{sec:experiments}
%======================================================================

We evaluate \method{} on four \baseline{} benchmarks~\citep{tttdiscover}, namely a step-function autocorrelation inequality (\textbf{AC1}, $1000$-element sequence, minimise $C_1$), the autocorrelation lower-bound problem (\textbf{AC2}, maximise $C_2$), circle packing of $n{=}26$ unit-square circles (\textbf{CP26}, maximise the radius sum) and the Erd\H{o}s minimum-overlap integral (\textbf{Erd\H{o}s}, minimise $C_5$). We ask whether \method{} discovers stronger solutions (\S\ref{sec:main_results}), whether preserved ensemble diversity drives those gains (\S\ref{sec:diversity}) and which components carry that mechanism (\S\ref{sec:ablation}).

\paragraph{Setup.} All runs use Qwen3-8B~\cite{qwen3} with $K{=}5$ rank-$16$ LoRA adapters, $\lambda_{\text{NNM}}{=}0.075$, $\alpha{=}0.1$, group size $G{=}8$ with $8$ groups per batch ($64$ rollouts per epoch), for $6$ epochs on one RTX~Pro~6000 (96\,GB; ${\approx}32$\,h per run), with \baseline{} prompts and verifiers unchanged. A rollout is correct when the verifier returns reward $>\!0$, and each correct rollout receives one algorithmic family label from the first matching rule in an ordered regular-expression list (otherwise \texttt{other}).

\begin{figure}[t]
\centering
\begin{subfigure}[t]{0.24\linewidth}
\centering
\includegraphics[width=\linewidth]{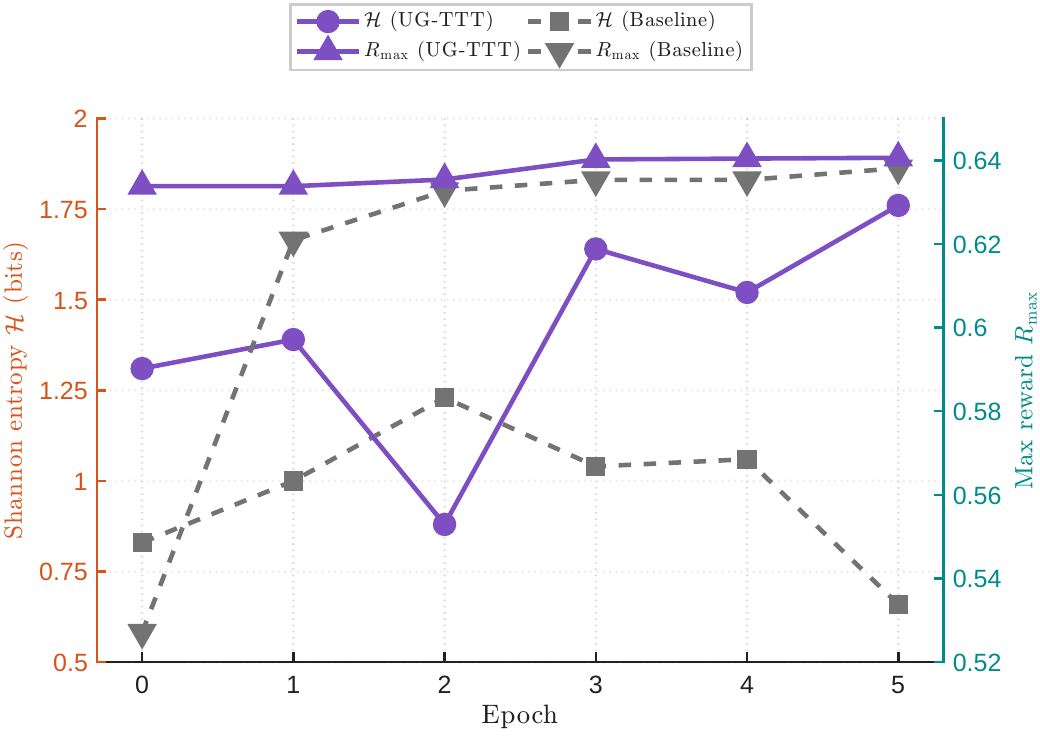}
\caption{AC1}
\label{fig:dyn_ac1}
\end{subfigure}%
\hfill
\begin{subfigure}[t]{0.24\linewidth}
\centering
\includegraphics[width=\linewidth]{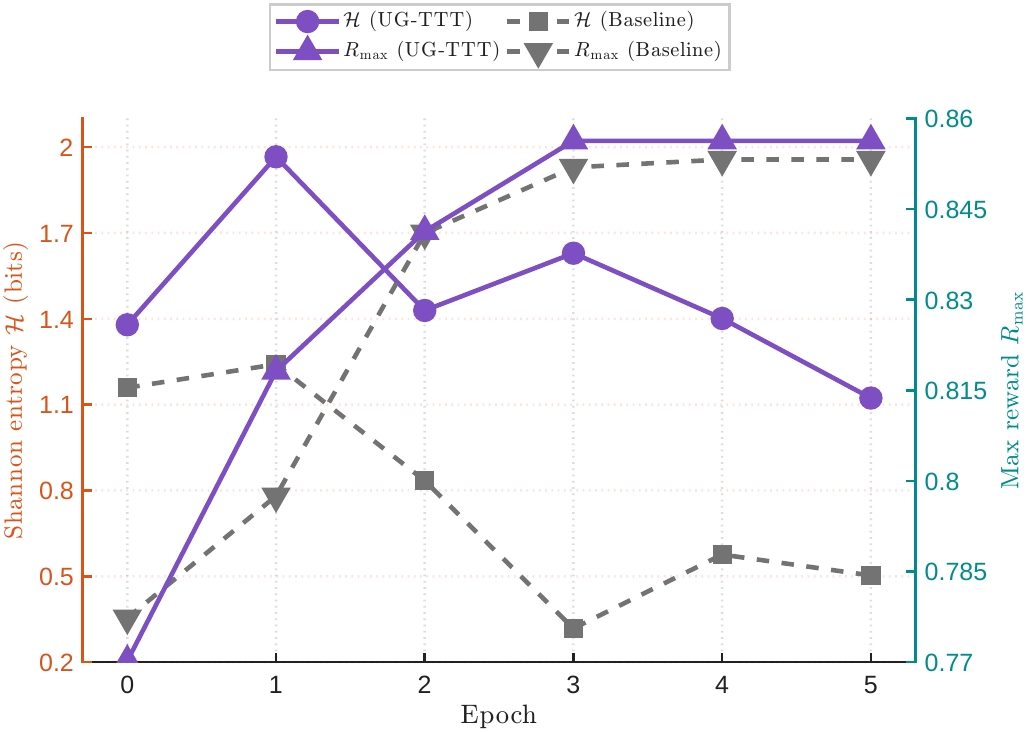}
\caption{AC2}
\label{fig:dyn_ac2}
\end{subfigure}%
\hfill
\begin{subfigure}[t]{0.24\linewidth}
\centering
\includegraphics[width=\linewidth]{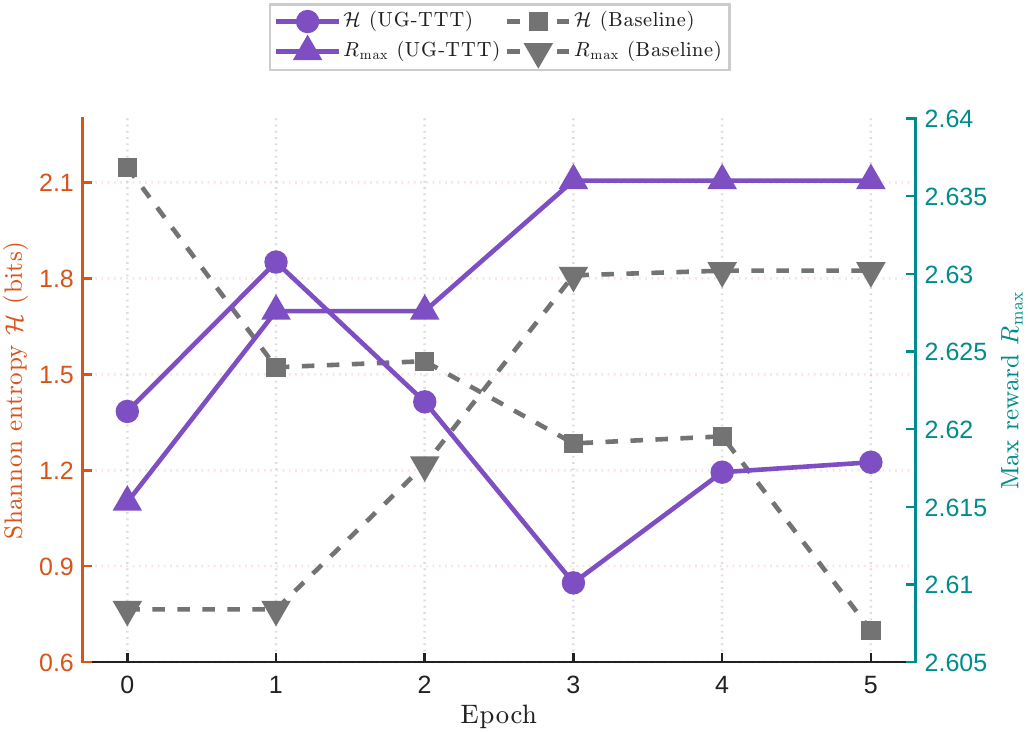}
\caption{CP26}
\label{fig:dyn_cp26}
\end{subfigure}%
\hfill
\begin{subfigure}[t]{0.24\linewidth}
\centering
\includegraphics[width=\linewidth]{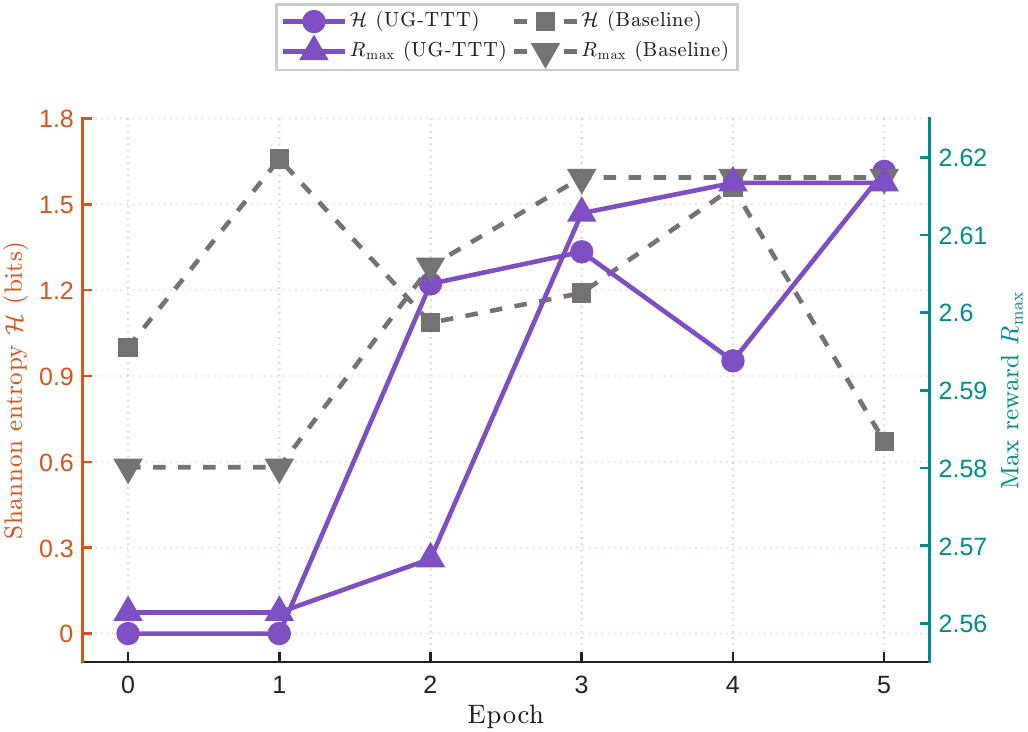}
\caption{Erd\H{o}s}
\label{fig:dyn_erdos}
\end{subfigure}
\caption{\textbf{Training dynamics across all four benchmarks.} Per-epoch Shannon entropy $H$ of the family distribution over correct rollouts (orange, left axis) and cumulative $R_{\max}$ (teal, right axis); dashed is baseline, solid is \method{}. The baseline contracts $H$ monotonically on every problem while \method{} sustains higher $H$ and matches or exceeds the baseline reward ceiling, so diversity preservation does not trade off against $R_{\max}$.}
\label{fig:training_dynamics}
\end{figure}
% --------------------------------------------------------------------------
\subsection{Main results}
\label{sec:main_results}
% --------------------------------------------------------------------------

Table~\ref{tab:main} reports training-time $R_{\max}$ and final-epoch family entropy $H$. \method{} improves $R_{\max}$ on three of four problems, matches the fourth to within $0.03\%$ of the bound ($\Delta\!=\!-0.0007$ on Erd\H{o}s), and retains $1.1$ to $1.7$ bits of $H$ in every case while the baseline collapses below $0.71$\,bits.

\begin{table}[t]
\centering
\small
\caption{\textbf{Main results across the four \baseline{} benchmarks.} Higher $R_{\max}$ is better; $R$ is $1/\text{upper bound}$ on AC1 and Erdős and a lower bound on AC2 and CP26. The CP26 ceiling matches the published \baseline{} state of the art, first reached at epoch 3.}
\label{tab:main}
\begin{tabular}{l >{\columncolor{tblBaseline}}r >{\columncolor{tblMethod}}r r >{\columncolor{tblBaseline}}r >{\columncolor{tblMethod}}r}
\toprule
\rowcolor{tblHeader}
        & \multicolumn{3}{c}{$R_{\max}$ during training} & \multicolumn{2}{c}{$H$ at epoch~5 (bits)} \\
\cmidrule(lr){2-4} \cmidrule(lr){5-6}
\rowcolor{tblHeader}
Problem & Baseline & \method{} & $\Delta$ & Baseline & \method{} \\
\midrule
AC1     & $0.6381$ & $\mathbf{0.6406}$ & $+0.0025$ & $0.70$ & $\mathbf{1.71}$ \\
AC2     & $0.8532$ & $\mathbf{0.8563}$ & $+0.0031$ & $0.50$ & $\mathbf{1.12}$ \\
CP26    & $2.6302$ & $\mathbf{2.6359}$ & $+0.0058$ & $0.70$ & $\mathbf{1.23}$ \\
Erd\H{o}s & $\mathbf{2.6174}$ & $2.6167$ & $-0.0007$ & $0.67$ & $\mathbf{1.62}$ \\
\bottomrule
\end{tabular}
\par\smallskip
{\footnotesize Streaming MI: enabled on AC1 and CP26; disabled on AC2 and Erd\H{o}s
(paired calibration in \S\ref{sec:ablation}, full diagnostic in
Appendix~\ref{app:streaming}).}
\end{table}

Absolute $\Delta$s mask where each gain sits on the discovery curve. On AC1, \method{} crosses $R\!\geq\!0.63$ at step~12 versus step~189 for the baseline (${\approx}16{\times}$ faster), and the baseline never reaches $0.64$ within budget. The CP26 $+0.0058$ matches the published \baseline{} ceiling of $2.6359$, which \method{} matches in $384$ rollouts ($6$ gradient steps $\times$ $8$ groups $\times$ $G{=}8$) versus the $25{,}600$ rollouts ($50$ steps of $512$) used by \baseline{} on the same Qwen3-8B model (\S4.1.2 and Tab.~9 of~\cite{Yuksekgonul2026Learning}). On AC2, the $+0.0031$ flip comes from the \texttt{differential\_evolution} family, which yields zero correct rollouts under the unregularised ensemble (Tab.~\ref{tab:idea}). Erd\H{o}s ties on the scalar bound, yet \method{} logs $13$ new-best events versus $7$ for the baseline ($+86\%$) and nearly doubles the discretisation resolution (mean $n_{\text{points}}\!=\!170$ vs.\ $89$), surfacing the Fourier initialisation revisited in \S\ref{sec:ablation}.

% --------------------------------------------------------------------------
\subsection{Diversity is the mechanism}
\label{sec:diversity}
% --------------------------------------------------------------------------

\begin{figure}[t]
\centering
\begin{subfigure}[t]{0.245\linewidth}
\centering
\includegraphics[width=\linewidth]{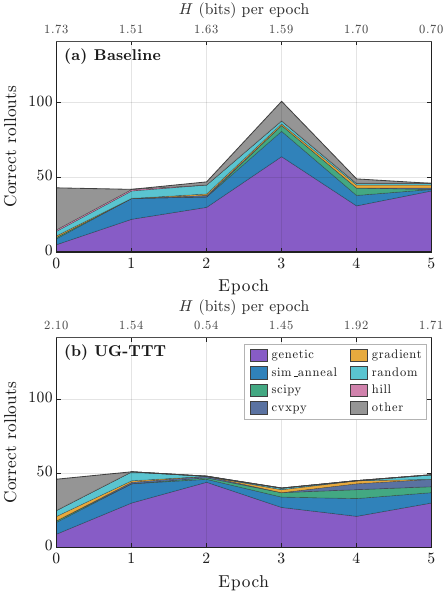}
\caption{AC1}
\label{fig:div_ac1}
\end{subfigure}%
\hfill
\begin{subfigure}[t]{0.245\linewidth}
\centering
\includegraphics[width=\linewidth]{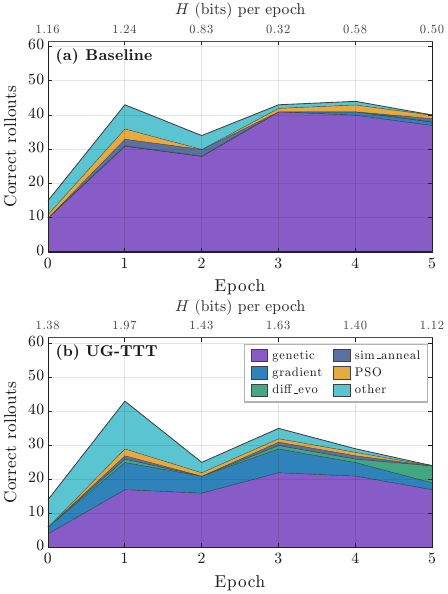}
\caption{AC2}
\label{fig:div_ac2}
\end{subfigure}%
\hfill
\begin{subfigure}[t]{0.245\linewidth}
\centering
\includegraphics[width=\linewidth]{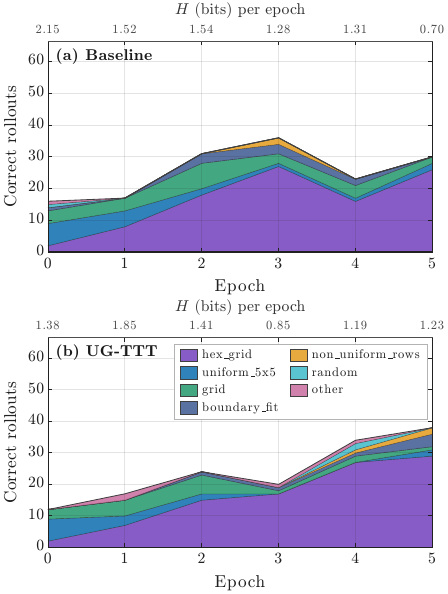}
\caption{CP26}
\label{fig:div_cp26}
\end{subfigure}%
\hfill
\begin{subfigure}[t]{0.245\linewidth}
\centering
\includegraphics[width=\linewidth]{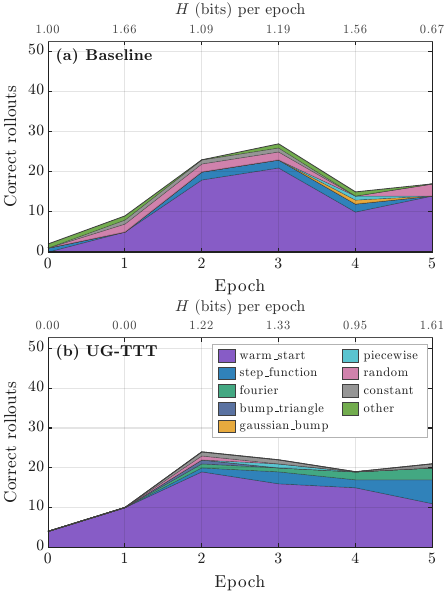}
\caption{Erd\H{o}s}
\label{fig:div_erdos}
\end{subfigure}
\caption{\textbf{Diversity collapse is universal and \method{} averts it.} Per-epoch family composition of correct rollouts on each benchmark; within each panel, top row is baseline (1 LoRA) and bottom is \method{} (5 LoRA + MI + NNM), with $H_{\mathrm{start}}\!\rightarrow\!H_{\mathrm{end}}$ reported above. Family vocabularies are problem-specific and identical across conditions, so the contrast lies in which families remain populated. Terminal $H\!=\!1.71/1.12/1.23/1.62$ bits under \method{} versus $0.70/0.50/0.70/0.67$ under the baseline.}
\label{fig:diversity_grid}
\end{figure}

Figure~\ref{fig:diversity_grid} tracks the per-epoch family mixture over correct rollouts on every problem. The pattern is uniform across the two family taxonomies (outer search loop on AC1/AC2, structural initialisation on CP26/Erd\H{o}s). The baseline narrows monotonically to $H\!\leq\!0.71$\,bits with one family at $82$ to $93\%$ of rollouts, whereas \method{} either holds a wider portfolio throughout (AC1, CP26) or opens across training (AC2, Erd\H{o}s) and finishes with $\geq 3$ live families on every problem. The collapse is therefore a universal failure mode of \baseline{}-style RL on discovery, and the diversity regulariser averts it. Mechanistically the $\beta$-coupling of Remark~1 (Eq.~\eqref{eq:gamma_eff}) drives the effect, since $\geff$ rises with reward-seeking pressure and sustains exploration on underrepresented families at the moment the baseline contracts. \S\ref{sec:ablation} confirms the link, as removing NNM drives MI to ${\sim}10^{-5}$ within two epochs and the family distribution narrows identically to the single-adapter baseline.

The families that survive under \method{} are exactly those that yield the winning solutions, namely the Cauchy-Schwarz target $\sqrt{2n}$ on AC1 ($18$/$30$ best rollouts versus $1$/$30$ for the baseline), the Fourier ansatz $h(x)\!=\!a+b\sin(kx)$ on Erd\H{o}s and \texttt{scipy.differential\_evolution} on AC2. Each motif sits inside a family the baseline extinguishes by the final epoch (Fig.~\ref{fig:diversity_grid}), so preserved diversity is a precondition for discovery. Full per-problem adoption rates appear in App.~\ref{app:idea}.

% --------------------------------------------------------------------------
\subsection{Ablation and computational cost}
\label{sec:ablation}
% --------------------------------------------------------------------------

Table~\ref{tab:ablation} ablates NNM against the full method on CP26 (the same $5$-LoRA comparator as Tab.~\ref{tab:main}). Without NNM, row spaces align within two epochs, ensemble MI falls to ${\sim}10^{-5}$ (3 to 4 orders of magnitude below the regularised run; Table~\ref{tab:ablation}), the MI bonus zeroes out, and the run reduces to unregularised $5$-LoRA RL; $R_{\max}$ stalls at $2.6302$, while the full method matches that ceiling on $25\%$ fewer training tokens ($2.04\!\times\!10^{6}$ versus $2.72\!\times\!10^{6}$). NNM is therefore the load-bearing component, keeping the ensemble informative enough for the MI bonus to deliver real exploration pressure to the policy gradient.
\begin{table}[t]
\centering
\small
\caption{\textbf{CP26 component ablation.} Removing NNM collapses ensemble MI within two epochs, zeroing the MI bonus and reverting the run to unregularised $5$-LoRA RL. Mean MI is per-rollout, averaged over epoch~5. ``Tokens to own $R_{\max}$'' is cumulative \texttt{rollout/num\_tokens} until the run first hits its own ceiling, while ``Tokens to $R\!=\!2.6302$'' is the cost of matching the no-NNM ceiling, giving a direct compute comparison at fixed reward.}
\label{tab:ablation}
\resizebox{\linewidth}{!}{%
\begin{tabular}{l c c r r}
\toprule
\rowcolor{tblHeader}
Configuration                                           & $R_{\max}$        & Mean MI (ep.\ 5) & Tokens to own $R_{\max}$ & Tokens to $R{=}2.6302$ \\
\midrule
\rowcolor{tblMethod}
Full \method{} (NNM + MI bonus)                         & $\mathbf{2.6359}$ & $\mathbf{2.2\!\times\!10^{-2}}$ & $\mathbf{2.33\!\times\!10^{6}}$ & $\mathbf{2.04\!\times\!10^{6}}$ \\
\rowcolor{tblBaseline}
No NNM (MI bonus collapses to $0$)                      & $2.6302$          & $6.6\!\times\!10^{-6}$           & $2.72\!\times\!10^{6}$ & $2.72\!\times\!10^{6}$ \\
\bottomrule
\end{tabular}%
}
\end{table}
\paragraph{Streaming MI, with paired calibration on AC2 and Erd\H{o}s.} Streaming early-stop (App.~\ref{app:streaming}) caps the thinking phase and re-checks ensemble MI before continuing, and we run it by default. Rollouts above the MI threshold continue past the cap to natural termination; only below-threshold rollouts are truncated, so the firing rate is the MI gate, not a hard length cut. It fires in $50.0\%$ of AC1 rollouts and $55.6\%$ of CP26 rollouts (saving $569$ and $757$ tokens, roughly $5\times$ the unregularised CP26 baseline rate of $10.7\%$), driving the $25\%$ token-budget reduction in Tab.~\ref{tab:ablation}; the Tab.~\ref{tab:main} $R_{\max}$ rollouts on AC1 and CP26 are themselves streamed. On AC2 and Erd\H{o}s, paired controls show streaming reducing $R_{\max}$ by $0.0148$ and $0.0017$ respectively, so we disable it there and report no-streaming numbers in Tab.~\ref{tab:main}. No length-versus-quality diagnostic predicts this asymmetry in advance, so the calibration is empirical. The Erd\H{o}s tie reflects \method{} matching the bound through a distinct family (Fourier basis with SLSQP at $n_{\text{points}}\!=\!400$) that the single-adapter baseline never enters.

\subsection*{Limitations} We evaluate on Qwen3-8B with $K\!=\!5$, one GPU configuration and a single seed across four benchmarks, so entropy gains and per-problem streaming behaviour may not transfer to other base models, larger $K$, or different initialisations. The ensemble incurs a roughly $K$-fold parameter overhead and a chunked $K$-way scoring pass, partially offset on AC1 and CP26 by streaming early-stop. The three coupled hyperparameters ($\alpha,\beta_{\mathrm{ref}},\gamma_{\max}$) and the fixed top-$7\%$ token threshold remain heuristic, and streaming calibration required paired pilots on AC2 and Erd\H{o}s. \method{} inherits a deterministic program-checker reward from \baseline{}, so porting to learned or noisy verifiers would require recalibrating the MI bonus.

\section{Background and related work}
\label{sec:related}
\textbf{RLHF and reward model optimization.} Reinforcement learning from human feedback (RLHF) has become the standard mechanism for aligning LLMs with human preferences \cite{Havrilla2024Teaching, Hong2025Cooper:, Rafailov2024Scaling}, producing gains in reasoning, commonsense, and code generation through Proximal Policy Optimization (PPO)-style optimization against a learned reward model \cite{2025DeepSeek-R1, Havrilla2024Teaching, Hong2025Cooper:}. Its central pathology is reward hacking: scalar rewards compress nuanced objectives into a single dimension, inherently suppressing exploration and enabling the exploitation of proxy loopholes while inducing overconfidence, hallucination, and distributional fragility \cite{Khalaf2025Inference-Time, Rafailov2024Scaling, Xia2025A, Shorinwa2024A, Farquhar2024Detecting, Tao2025Revisiting}.

\textbf{Test-time training.} Test-time training (TTT) adapts model parameters at inference using self-supervised signals or retrieved neighbors \cite{Hu2025Test-Time, Hardt2023Test-Time, Akyurek2024The}, often with parameter-efficient adapters to mitigate forgetting \cite{Hu2025Test-Time, Moradi2025Continuous}, and has delivered substantial gains on the Abstraction and Reasoning Corpus (ARC) and BIG-Bench Hard \cite{Akyurek2024The}. Like RLHF, it inherits mode collapse under sparse outcome rewards \cite{Thomas2025Test-Time}: without dense intrinsic signals, adaptation converges prematurely on familiar solution paths rather than advancing toward the knowledge frontier.

\textbf{Uncertainty quantification.} Foundational work on uncertainty quantification (UQ) separates aleatoric from epistemic uncertainty and proposes Bayesian networks, deep ensembles, Monte Carlo dropout, and evidential methods as estimation strategies \cite{Abdar2020A, Charpentier2022Disentangling, Hullermeier2019Aleatoric, Caldeira2020Deeply}. In LLMs, UQ has been deployed for hallucination detection through token probabilities, verbalized confidence, ensemble disagreement, and conformal intervals \cite{Xia2025A, Shorinwa2024A, Farquhar2024Detecting, Tao2025Revisiting, Liu2025Uncertainty, Vashurin2024Benchmarking, Huang2023Look}, with production systems such as LM-Polygraph \cite{Fadeeva2023LM-Polygraph:} and entropy-based confabulation detectors \cite{Farquhar2024Detecting}. These methods remain post-hoc: they flag unreliable outputs but do not integrate into the training signal, and most fail to cleanly separate epistemic from aleatoric sources at LLM scale \cite{Liu2025Uncertainty, Abdar2020A}.

\textbf{Intrinsic motivation and MI-based exploration.} Curiosity bonuses and entropy regularization have been proposed to counter exploitation bias in RL with LLMs \cite{Du2023Guiding, Liao2025Enhancing, Gao2025Navigate, Dwaracherla2024Efficient, Levy2025WorldLLM:, Cheng2025Reasoning}, though classical novelty rewards suffer vanishing incentives in large action spaces \cite{Dai2025CDE:, Yuan2022Renyi}. Recent work uses ensemble disagreement, including LoRA-based variants, as an intrinsic signal tied to the knowledge boundary \cite{Shorinwa2024A, Hullermeier2019Aleatoric}, with actor-wise perplexity bonuses and critic-wise value variance integrated into advantage functions \cite{Dai2025CDE:}. Frameworks such as Curiosity-Driven Exploration (CDE) and i-MENTOR demonstrate that maintained entropy prevents premature convergence in sparse-reward reinforcement learning with verifiable rewards (RLVR) \cite{Dai2025CDE:, Gao2025Navigate}, but balancing intrinsic against outcome rewards risks calibration collapse \cite{Dai2025CDE:}. None of these approaches target epistemic uncertainty at the granularity required for frontier exploration, which is the gap \method{} addresses.
%======================================================================
\section{Conclusion}
\label{sec:conclusion}
%======================================================================

We argue that diversity collapse in test-time discovery systems is a structural consequence of single-model, scalar-reward optimization: policy gradients suppress high-variance candidates, and a single network cannot disentangle epistemic from aleatoric uncertainty. \method{} addresses both by introducing an ensemble of low-rank adapters that provides a token-level mutual information signal, combined with a nuclear norm regularizer to maintain subspace diversity and a temperature-coupled exploration bonus.

\method{} improves maximum reward on three of four benchmarks and matches the fourth while discovering qualitatively distinct solutions, sustaining substantially higher mutual information and late-stage entropy. Ablations identify the nuclear norm regularizer as critical, with its removal causing rapid collapse of the uncertainty signal and reversion to standard reinforcement learning behavior.

\section*{LLM usage}
\vspace{-0.25em}
\texttt{Sonnet 4.6} by \texttt{Anthropic} was used for editing, proofreading, and resolving \LaTeX~formatting issues in this document. \texttt{Claude Code} by \texttt{Anthropic} was used to assist with code writing in the implementation of \texttt{\method{}}.

\bibliographystyle{plainnat}
\bibliography{references}

\newpage
\section*{Appendix}
\renewcommand{\arraystretch}{1}
\appendix

\section{Baseline objective: \baseline{} in full notation}
\label{app:baseline}

We collect here, for reference, the \baseline{} training signal that \method{} builds on. \baseline{} optimises a state-conditional entropic objective with an adaptive temperature $\beta(s)$,
\begin{equation}
    J_\beta(\theta) \;=\; \mathbb{E}_{\,s\sim\mathrm{reuse}(\mathcal{H})}\!\Bigg[\,\frac{1}{\beta(s)}\log\,\mathbb{E}_{\,a\sim\pi_\theta(\cdot\mid q,s)}\!\bigl[e^{\beta(s)\,\Rexec(s,a)}\bigr]\Bigg],
    \label{eq:baseline_objective}
\end{equation}
which interpolates between expected reward ($\beta\!\to\!0$) and the per-state maximum ($\beta\!\to\!\infty$). The outer expectation is over parent states $s$ drawn from the reuse distribution over the PUCT history $\mathcal{H}$, and the inner expectation is over rollouts $a$ sampled from the policy at prompt $q$ and parent $s$. For every group, $\beta(s)$ is selected as the unique $\beta>0$ such that the softmax $q_\beta(a)\!\propto\!\exp(\beta \Rexec(a))$ has a fixed $\ln 2$ KL budget against uniform,
\begin{equation}
    \mathrm{KL}\!\bigl(q_\beta\,\bigl\|\,\mathrm{Unif}\bigr)\;=\;\ln 2,
    \label{eq:beta_root}
\end{equation}
which exists and is unique whenever the within-group rewards are not all identical, since $\mathrm{KL}(q_\beta\|\mathrm{Unif})$ is continuous and strictly increasing in $\beta$ on $(0,\infty)$ with limits $0$ and $\ln G$. We solve Eq.~\eqref{eq:beta_root} by 60 iterations of bisection on $[0,10^{6}]$; on the degenerate equal-reward set, the bisection saturates at the upper bracket and we fall back to $A_i\equiv 0$. Given $\beta(s)$, rollout $i\!\in\!\{1,\ldots,G\}$ receives the leave-one-out entropic weight and advantage
\begin{equation}
    w_i \;=\; \frac{\exp\!\bigl(\beta\,\Rexec^{(i)}\bigr)}{\tfrac{1}{G-1}\sum_{j\neq i}\exp\!\bigl(\beta\,\Rexec^{(j)}\bigr)},
    \qquad A_i \;=\; w_i - 1,
    \label{eq:loo_adv}
\end{equation}
which centres the advantage within each group without a learned value function. The rollouts are trained with the clipped importance-sampling loss
\begin{equation}
    \mathcal{L}_{\mathrm{PG}}(\theta)\;=\;-\sum_{t=1}^{|o|}\!\min\!\Bigl[\,\rho_t(\theta)\,A_i,\;\mathrm{clip}\bigl(\rho_t(\theta),\,1{-}\epsilon,\,1{+}\epsilon\bigr)\,A_i\Bigr],
    \label{eq:pg_loss}
\end{equation}
with likelihood ratio $\rho_t(\theta)=\pi_\theta(o_t\mid q,o_{<t})/\pi_{\theta_{\mathrm{old}}}(o_t\mid q,o_{<t})$, reducing to plain importance sampling when $\epsilon$ is disabled. A KL-to-base penalty $\lambda_{\mathrm{KL}}\,\mathrm{KL}(\pi_\theta\|\pi_{\mathrm{base}})$ is added when $\lambda_{\mathrm{KL}}\!>\!0$.

The per-adapter form used inside $\method{}$ replaces $A_i$ with $A_i^{\mathrm{shaped}}$ (Eq.~\eqref{eq:a_shaped}):
\begin{equation}
    \mathcal{L}_{\mathrm{PG}}^{(k)}(\theta_k)\;=\;-\!\sum_{t=1}^{|o|}\!\min\!\Bigl[\,\rho_t^{(k)}(\theta_k)\,A_i^{\mathrm{shaped}},\;\mathrm{clip}\bigl(\rho_t^{(k)},\,1{-}\epsilon,\,1{+}\epsilon\bigr)\,A_i^{\mathrm{shaped}}\Bigr],
    \label{eq:pg_loss_k}
\end{equation}
with per-adapter ratio $\rho_t^{(k)}(\theta_k)=\pi_{\theta_k}(o_t\mid q,o_{<t})/\pi_{\theta_{k,\mathrm{old}}}(o_t\mid q,o_{<t})$.

\paragraph{Token-wise KL adjustment.}\label{app:kl_adjust}
When $\lambda_{\mathrm{KL}}>0$, the scalar $A_i^{\mathrm{shaped}}$ is broadcast to a per-token tensor and the token-wise correction of~\citet{tttdiscover} is applied before the clipped-IS reduction. Concretely, for each position $t$ the advantage entry is adjusted by $-\lambda_{\mathrm{KL}}\,\bigl(\log\pi_{\theta}(o_t\mid q,o_{<t}) - \log\pi_{\mathrm{base}}(o_t\mid q,o_{<t})\bigr)$, which under sampling from $\pi_\theta$ gives the standard k1 single-sample stochastic estimator of $\nabla_\theta\,\lambda_{\mathrm{KL}}\mathrm{KL}(\pi_\theta\|\pi_{\mathrm{base}})$ folded into the per-token PG target---no second autograd pass through the policy is needed. The transcription is identical to the reference implementation and is stated here only to fix notation used in the proof of Proposition~1.

\section{Full training algorithm}
\label{app:algorithm}

\begin{algorithm}[H]
\caption{\method{} training step (one group of $G$ rollouts).}
\label{alg:ttt_mi}
\begin{algorithmic}[1]
\Require Ensemble $\{\theta_k\}_{k=1}^K$; parent state $s$ from PUCT buffer; group size $G$.
\Require Coefficients $\alpha$, $\beta_{\mathrm{ref}}$, $\gamma_{\max}$, $\lambda_{\mathrm{KL}}$, $\lambda_{\mathrm{NNM}}$.
\Ensure Updated ensemble $\{\theta_k\}_{k=1}^K$.
\vspace{2pt}
\State \textbf{// Stage 1: Rollout generation (round-robin over adapters)}
\State Build prompt $q$ from $(d, s)$.
\For{$i = 1, \ldots, G$}
    \State $k_i \gets \bigl((i - 1) \bmod K\bigr) + 1$ \hfill $\triangleright$ 1-indexed; $K\mid G$ so the assignment pattern $(1,2,\ldots,K,1,2,\ldots)$ repeats identically every group
    \State $o_i \sim \pi_{\theta_{k_i}}(\cdot \mid q, s)$
    \State $R_{\mathrm{exec}}^{(i)} \gets \textsc{ProgramChecker}(s, o_i)$
\EndFor
\vspace{2pt}
\State \textbf{// Stage 2: Mutual-information scoring (single GPU pass)}
\State $\{p_{\theta_k}\}_{k=1}^{K} \gets \textsc{ChunkedScoreAllAdapters}(o_{1:G})$
\For{$i = 1, \ldots, G$}
    \For{$t = 1, \ldots, |o_i|$}
        \State $\bar{p}_t \gets \tfrac{1}{K}\sum_{k=1}^{K} p_{\theta_k}(\cdot \mid q, o_{<t}^{(i)})$
        \State $\MI_t^{(i)} \gets H(\bar{p}_t) - \tfrac{1}{K}\sum_{k=1}^{K} H\bigl(p_{\theta_k}(\cdot \mid q, o_{<t}^{(i)})\bigr)$
        \Statex \hfill $\triangleright$ Eq.~\eqref{eq:true_mi}
    \EndFor
    \State $U_i \gets \operatorname*{top\text{-}7\%\text{-}mean}_{t} \MI_t^{(i)}$ \hfill $\triangleright$ Eq.~\eqref{eq:top7}
\EndFor
\vspace{2pt}
\State \textbf{// Stage 3: Adaptive temperature and LOO advantage (on $\Rexec$ only)}
\State $\beta(s) \gets \textsc{BisectionSolve}\bigl\{\beta>0 : \mathrm{KL}(q_\beta \,\|\, \mathrm{Unif}) = \ln 2\bigr\}$ on $\{R_{\mathrm{exec}}^{(i)}\}$
\Statex \hfill $\triangleright$ Eq.~\eqref{eq:beta_root}
\For{$i = 1, \ldots, G$}
    \State $w_i \gets \exp(\beta R_{\mathrm{exec}}^{(i)}) \big/ \tfrac{1}{G-1}\sum_{j \neq i} \exp(\beta R_{\mathrm{exec}}^{(j)})$
    \State $A_i \gets w_i - 1$ \hfill $\triangleright$ Eq.~\eqref{eq:loo_adv}
\EndFor
\vspace{2pt}
\State \textbf{// Stage 4: MI-shaped advantage}
\State $\overline{U} \gets \tfrac{1}{G}\sum_i U_i$; \ \ $\sigma_U \gets \mathrm{std}_{\text{unbiased}}(U)$ (cf.\ Eq.~\eqref{eq:mi_standardise}); \ \ $\overline{|A|} \gets \tfrac{1}{G}\sum_j |A_j|$
\State $\geff(s) \gets \alpha \cdot \min\!\bigl(\beta(s)/\beta_{\mathrm{ref}},\; \gamma_{\max}\bigr)$ \hfill $\triangleright$ Eq.~\eqref{eq:gamma_eff}
\For{$i = 1, \ldots, G$}
    \State $\tilde U_i \gets \mathrm{clip}\!\bigl((U_i - \overline{U})/\sigma_U,\; -3,\; 3\bigr)$ \hfill $\triangleright$ Eq.~\eqref{eq:mi_standardise}
    \State $A_i^{\mathrm{shaped}} \gets A_i + \geff(s) \cdot \overline{|A|} \cdot \tilde U_i$ \hfill $\triangleright$ Eq.~\eqref{eq:a_shaped}
\EndFor
\If{$\lambda_{\mathrm{KL}} > 0$}
    \State Broadcast $A_i^{\mathrm{shaped}}$ per-token; add token-wise KL correction against $\pi_{\mathrm{base}}$ (Appendix~\ref{app:kl_adjust})
\EndIf
\vspace{2pt}
\State \textbf{// Stage 5: Per-adapter backward, NNM backward, then optimiser step}
\For{$k = 1, \ldots, K$}
    \State Compute $\tfrac{1}{K}\,\mathcal{L}_{\mathrm{PG}}^{(k)}(\theta_k)$ over all $G$ rollouts and call \texttt{backward()}; gradients accumulate into $\theta_k.\mathrm{grad}$ \hfill $\triangleright$ Eq.~\eqref{eq:pg_loss_k}
    \State Release adapter $k$'s computation graph before advancing
\EndFor
\Statex \hspace{1.2em}\textit{// KL-to-base contribution is already inside $\mathcal{L}_{\mathrm{PG}}^{(k)}$ via the Stage~4 token-wise correction (App.~\ref{app:kl_adjust})}
\State Compute $\lambda_{\mathrm{NNM}}\,\mathcal{L}_{\mathrm{NNM}}$ and call \texttt{backward()}; gradients accumulate into all $\{\theta_k\}$ \hfill $\triangleright$ Eq.~\eqref{eq:Wstack}
\For{$k = 1, \ldots, K$}
    \State AdamW step on $\theta_k$ using its accumulated gradient; zero $\theta_k.\mathrm{grad}$
\EndFor
\end{algorithmic}
\end{algorithm}

\section{Proofs}
\label{app:proofs}

\subsection{Proof of Proposition~1 (bounded scale perturbation of the shaped advantage)}

Fix a parent state $s$ and its group of $G$ rollouts. If $\sigma_U\!=\!0$, Eq.~\eqref{eq:mi_standardise} sets $\tilde U_i\equiv 0$ by convention, and $A_i^{\mathrm{shaped}}=A_i$ identically. Otherwise
\begin{equation*}
    \sum_{i=1}^{G}\tilde U_i \;=\; \sum_{i=1}^{G}\mathrm{clip}\!\left(\frac{U_i-\overline U}{\sigma_U},\,-3,\,3\right).
\end{equation*}
Let $z_i:=(U_i-\overline U)/\sigma_U$ and $\mathcal{C}=\{i:|z_i|>3\}$. By centring, $\sum_{i=1}^G z_i = 0$, so
\begin{equation*}
    \sum_{i=1}^G \tilde U_i \;=\; \sum_{i\notin\mathcal{C}} z_i \;+\; \sum_{i\in\mathcal{C}} \mathrm{sign}(z_i)\cdot 3
    \;=\; -\sum_{i\in\mathcal{C}} z_i \;+\; \sum_{i\in\mathcal{C}}\mathrm{sign}(z_i)\cdot 3
    \;=\; \sum_{i\in\mathcal{C}}\bigl(\mathrm{sign}(z_i)\cdot 3 - z_i\bigr).
\end{equation*}
For $i\in\mathcal{C}$ we have $|z_i|>3$, and a direct case split on $\mathrm{sign}(z_i)$ gives $|\mathrm{sign}(z_i)\cdot 3 - z_i| = |z_i|-3$, so by the triangle inequality
\begin{equation*}
    \Bigl|\sum_{i=1}^G \tilde U_i\Bigr| \;\le\; \sum_{i\in\mathcal{C}}\bigl(|z_i|-3\bigr).
\end{equation*}
By Cauchy--Schwarz, $\sum_{i\in\mathcal{C}}|z_i| \le \sqrt{|\mathcal{C}|\,\sum_i z_i^2} = \sqrt{|\mathcal{C}|(G-1)}$ (using $\sigma_U$ unbiased, so $\sum_i z_i^2=G-1$); a looser but more interpretable bound uses $|z_i|\le\sqrt{G-1}$ to give $\sum_{i\in\mathcal{C}}\bigl(|z_i|-3\bigr) \le |\mathcal{C}|(\sqrt{G-1}-3)$. When $\mathcal{C}=\varnothing$ both sides are exactly $0$; note in particular that $|z_i|\le\sqrt{G-1}$ forces $\mathcal{C}=\varnothing$ whenever $G\le 10$, so the looser bound is non-vacuous only in the regime $G\ge 11$ (where $\sqrt{G-1}-3>0$). Substituting into Eq.~\eqref{eq:a_shaped} and using that neither $\geff(s)$ nor $\overline{|A|}$ depends on $i$,
\begin{equation*}
    \biggl|\sum_{i=1}^G A_i^{\mathrm{shaped}} - \sum_{i=1}^G A_i\biggr| \;=\; \geff(s)\,\overline{|A|}\,\biggl|\sum_{i=1}^G \tilde U_i\biggr|
    \;\le\; \geff(s)\,\overline{|A|}\,|\mathcal{C}|\bigl(\sqrt{G-1}-3\bigr),
\end{equation*}
which is the statement of the proposition.

We emphasise that this is strictly weaker than unbiasedness of the policy-gradient estimator $\sum_i A_i\,\nabla\log\pi(o_i)$ in the standard sense~\citep{advshaping}: because $\tilde U_i$ depends on rollout $i$ through $U_i$, the substitution $A_i\!\mapsto\!A_i^{\mathrm{shaped}}$ does not preserve the expected gradient. The bonus is deliberately biased; the directional bias toward high-MI rollouts is the exploration mechanism. Proposition~1 isolates the bias to the gradient direction by bounding the residual perturbation to the aggregate within-group magnitude. \hfill$\square$

\subsection{Proof of Proposition~2 (the global maximum is subspace-orthogonal)}

Fix a layer $\ell$ and write $A_k:=A_\ell^{(k)}\in\mathbb{R}^{r\times d_{\mathrm{in}}}$. The problem is
\begin{equation*}
    \max_{A_1,\ldots,A_K}\; \bigl\|W\bigr\|_*,\qquad W = \bigl[A_1; \ldots; A_K\bigr]\in\mathbb{R}^{Kr\times d_{\mathrm{in}}},\qquad \|A_k\|_F = c\ \forall k.
\end{equation*}
Let $\sigma_1,\ldots,\sigma_{m}$ denote the nonzero singular values of $W$, where $m=\mathrm{rank}(W)\le \min(Kr,d_{\mathrm{in}})=Kr$ (using the hypothesis $Kr\le d_{\mathrm{in}}$). By Cauchy--Schwarz applied to $(\sigma_1,\ldots,\sigma_m)$ and the all-ones vector,
\begin{equation*}
    \|W\|_*^2 \;=\; \Bigl(\sum_{i=1}^m \sigma_i\Bigr)^{\!2} \;\le\; m \sum_{i=1}^m \sigma_i^2 \;\le\; Kr \cdot \|W\|_F^2 \;=\; Kr \sum_{k=1}^K c^2 \;=\; K^2 r\, c^2,
\end{equation*}
with equality in both inequalities iff $m = Kr$ (i.e.\ $W$ has full row rank, which requires and is feasible exactly when $Kr\le d_{\mathrm{in}}$, the standing hypothesis) and all $Kr$ singular values are equal. Let $s$ denote this common singular value; then $\|W\|_F^2 = Kr\,s^2 = Kc^2$ gives $s = c/\sqrt{r}$. Under the full-row-rank hypothesis just established, equal singular values are equivalent to $WW^\top = s^2 I_{Kr}$, i.e.\ the rows of $W$ are mutually orthogonal with common norm $s$. Restricted to a single block $A_k$, this forces $A_k$ to have $r$ equal singular values all equal to $s = c/\sqrt{r}$. Across blocks, mutual row orthogonality of $W$ forces the row spaces of $A_1,\ldots,A_K$ to be mutually orthogonal subspaces of $\mathbb{R}^{d_{\mathrm{in}}}$; this is feasible iff $Kr\le d_{\mathrm{in}}$, which holds by hypothesis. The global maximum of $\|W\|_*$ on the constraint set therefore satisfies both conditions, as claimed. \hfill$\square$

\subsection{Proof of Proposition~3 (\baseline{} as a special case)}

Under either branch the ensemble's mixture $\bar p_t$ in Eq.~\eqref{eq:true_mi} coincides with each member's distribution: in (a) trivially because $K=1$, and in (b) because identical weights yield identical $p_{\theta_k}$. Each term of the decomposition $H(\bar p_t) - \tfrac{1}{K}\sum_k H(p_{\theta_k})$ then equals the same quantity, so $\MI_t\equiv 0$. Consequently $U_i\equiv 0$, $\overline U=0$, $\sigma_U=0$, and by the convention stated below Eq.~\eqref{eq:mi_standardise} we set $\tilde U_i\equiv 0$. The shaped advantage reduces to $A_i^{\mathrm{shaped}}=A_i$ and Eq.~\eqref{eq:pg_loss_k} reduces to Eq.~\eqref{eq:pg_loss}. The hypothesis $\lambda_{\mathrm{NNM}}=0$ removes $\mathcal{L}_{\mathrm{NNM}}$ from Eq.~\eqref{eq:total_loss} outright (this assumption is required even in branch (a), since for $K=1$ the regulariser $-\|A^{(1)}_\ell\|_*$ is a non-trivial function of $\theta_1$ that would otherwise perturb the policy update). What remains of Eq.~\eqref{eq:total_loss} is $\mathcal{L}_{\mathrm{PG}}(\theta_1) + \lambda_{\mathrm{KL}}\mathrm{KL}(\pi_\theta\|\pi_{\mathrm{base}})$, which is the \baseline{} objective of~\citet{tttdiscover}. In branch (b), identical-init plus shared-data training (Sec.~\ref{sec:method_loss}) keeps the $K$ adapters tied at every step, so the per-adapter update is byte-identical to baseline's single-adapter update. \hfill$\square$

\section{Implementation notes}
\label{app:implementation}

\paragraph{Chunked MI and base-logprob scoring.}
Two distinct logit tensors arise during scoring. (i) The \emph{ensemble MI} path materialises an $(K,\,\mathrm{chunk},\,V)$ fp32 logit tensor inside the chunked LM head; for Qwen3-8B with $K\!=\!5$, $V\!=\!152{,}064$, and $\mathrm{chunk}\!=\!256$ this peaks at $\approx\!0.78$\,GB per chunk. The MI decomposition of Eq.~\eqref{eq:true_mi} is summed into a running per-rollout buffer that reuses the same chunk's $\log\!\mathrm{softmax}$ (already computed for target-token gather), so the MI contribution costs $<\!0.1\%$ of the LM-head matmul. (ii) The \emph{base-model logprob} path needed for the per-token KL-to-base correction (App.~\ref{app:kl_adjust}) materialises a $(1,T,V)$ fp32 logit tensor; for $T\!=\!26$k this would peak at $\approx\!16$\,GB without chunking, and we reduce it to $\approx\!8.1$\,GB by streaming $\log\!\mathrm{softmax}$ over the same $\mathrm{chunk}\!=\!256$ granularity and releasing each chunk's allocation before the next is loaded.

\paragraph{Top-$7\%$ threshold.}
\citet{wang80_20} shows that the RL signal in LLMs should concentrate on the high-entropy minority of tokens, the pivotal decision points, rather than be diluted across boilerplate. The top-$7\%$ choice is motivated in §\ref{sec:method_mi}; it adapts to sequence length so an $n$-token response contributes $\lceil 0.07n\rceil$ positions, which remains well-calibrated across the 7k--15k-token range typical of reasoning outputs.

\paragraph{Sequential per-adapter backward.}
A single joint $K$-adapter forward--backward retains $K$ parallel computation graphs and scales as $\mathcal{O}(K\!\cdot\!T\!\cdot\!L\!\cdot\!D)$ activation memory, exceeding 80\,GB on our setup. We therefore iterate over adapters in Stage~5 of Algorithm~\ref{alg:ttt_mi} (the mLoRA scheme cited in §\ref{sec:method_loss}): each adapter's forward and backward are computed in turn, the computation graph is freed before the next adapter's forward, and gradients are accumulated into a shared buffer for the final AdamW step.

\paragraph{Restricting regularisation to $A^{(k)}$.}
In pilot runs, regularising both $A^{(k)}$ and $B^{(k)}$ produced oscillatory optimisation: a penalty applied to $B^{(k)}$ was absorbed by a compensating rescaling of $A^{(k)}$, and vice versa, giving no net increase in subspace diversity while destabilising the RL loss. This matches the empirical observation of~\citet{uprlhf}. Leaving $B^{(k)}$ free removes the degeneracy.

\section{Hyperparameters}
\label{app:hyperparams}

Tables~\ref{tab:hp_shared} and~\ref{tab:hp_method} list every hyperparameter used in all reported runs. Values are identical across problems unless noted; per-problem deviations are listed in the notes column.

\begin{table}[H]
\centering
\small
\caption{\textbf{Shared hyperparameters} (identical for \baseline{} and \method{} on all four problems).}
\label{tab:hp_shared}
\renewcommand{\arraystretch}{1.2}
\setlength{\tabcolsep}{6pt}
\begin{tabular}{@{}lll@{}}
\toprule
\rowcolor{tblHeader}
Hyperparameter & Value & Notes \\
\midrule
\multicolumn{3}{@{}l}{\textit{Model}} \\
Base model & Qwen3-8B & fp16 precision \\
LoRA target modules & q, k, v, o projections & all attention projections \\
LoRA rank $r$ & 16 & \\
LoRA alpha & 32 & effective scale $= \alpha / r = 2$ \\
LoRA dropout & 0.05 & \\
\midrule
\multicolumn{3}{@{}l}{\textit{Optimiser}} \\
Optimiser & AdamW & \\
Learning rate & $4 \times 10^{-5}$ & \\
\midrule
\multicolumn{3}{@{}l}{\textit{RL training}} \\
Training epochs & 6 & \\
Group size $G$ & 8 & rollouts per PUCT node \\
Groups per batch & 8 & gradient accumulation \\
Rollout temperature & 1.0 & \\
Advantage estimator & entropic adaptive-$\beta$ & Eq.~\eqref{eq:baseline_objective} \\
$\beta_{\mathrm{ref}}$ & 2.0 & bisection target for KL$\,{=}\,\ln 2$ \\
IS clip $\epsilon$ & 0.2 & \\
KL penalty $\lambda_{\mathrm{KL}}$ & 0.01 & \\
Remove constant-reward groups & yes & \\
\midrule
\multicolumn{3}{@{}l}{\textit{Sampling and context}} \\
Sampler & PUCT backprop & \\
Context window & 32\,768 tokens & Qwen3-8B native \\
Phase-1 (thinking) token cap & 26\,000 & hard cap on thinking tokens \\
Phase-2 wrap budget & 6\,000 tokens & code phase; streaming runs only \\
\bottomrule
\end{tabular}
\end{table}

\begin{table}[H]
\centering
\small
\caption{\textbf{\method{}-specific hyperparameters.} The baseline sets $K{=}1$, $\alpha{=}0$, $\lambda_{\mathrm{NNM}}{=}0$, and disables streaming MI.}
\label{tab:hp_method}
\renewcommand{\arraystretch}{1.2}
\setlength{\tabcolsep}{6pt}
\begin{tabular}{@{}lll@{}}
\toprule
\rowcolor{tblHeader}
Hyperparameter & Value & Notes \\
\midrule
\multicolumn{3}{@{}l}{\textit{Ensemble}} \\
Number of adapters $K$ & 5 & round-robin rollout assignment \\
\midrule
\multicolumn{3}{@{}l}{\textit{MI exploration bonus (Sec.~\ref{sec:method_mi_shaping})}} \\
MI metric & true MI, top-7\% tokens & Eq.~\eqref{eq:true_mi}, Eq.~\eqref{eq:top7} \\
MI coefficient $\alpha$ & 0.1 & base exploration strength \\
$\gamma_{\max}$ ratio & 10.0 & clips $\gamma_{\mathrm{eff}} / \alpha$ \\
MI clip range & $[-3, 3]$ & on standardised $\tilde{U}_i$ \\
\midrule
\multicolumn{3}{@{}l}{\textit{Nuclear-norm regularisation (Sec.~\ref{sec:method_nnm})}} \\
NNM coefficient $\lambda_{\mathrm{NNM}}$ & 0.075 & applied to $A^{(k)}$ matrices only \\
Frobenius-normalised variant & no & raw nuclear norm used \\
\midrule
\multicolumn{3}{@{}l}{\textit{Streaming MI early-stop (Appendix~\ref{app:streaming})}} \\
Enabled & AC1, CP26 & disabled on AC2 and Erd\H{o}s \\
Window size $W$ & 4\,096 tokens & \\
Check interval $C$ & 2\,048 tokens & \\
Min generation before check & 4\,096 tokens & \\
MI threshold percentile & 25th & of historical window MI \\
Warmup epochs & 3 & no streaming before epoch 3 \\
\bottomrule
\end{tabular}
\end{table}

\section{Streaming MI early-stop and length-reward diagnostic}
\label{app:streaming}

\paragraph{Mechanism as logged.}
The streaming early-stop is implemented as a hard cap on the thinking
phase: when a rollout reaches $4096$ thinking tokens (occasionally
$6144$ on Erd\H{o}s where the cap was widened), the windowed ensemble
MI is compared to the running $25^{\text{th}}$ percentile of historical
MI; if the rollout is below threshold, the thinking phase is closed and
the rollout is forced into a bounded code phase. The MI check is
evaluated once, at the cap: rollouts below the historical
$25^{\text{th}}$ percentile are truncated to a bounded code phase,
while rollouts above threshold continue past the cap to natural
termination. The recorded \texttt{streaming\_mi\_stop\_step} therefore
coincides with the cap (it is the step at which the gate is queried),
but the gate itself is the MI threshold, not the cap; the
$50.0\%/55.6\%$ firing rates on AC1 and CP26 are the rates at which the
MI check chose to truncate, not a fixed length cut. Non-streamed
rollouts within the same run either emitted \texttt{</think>} before
the cap or were above threshold at the cap and continued to natural
termination.

\paragraph{Per-problem firing and savings.}
\begin{center}
\small
\begin{tabular}{l c c c c}
\toprule
\rowcolor{tblHeader}
Problem & Firing rate & Mean tokens saved per rollout & Mean tokens saved per firing & Streaming setting in Table~\ref{tab:main} \\
\midrule
AC1   & $50.0\%$  & $569$ & $1{,}138$ & enabled \\
CP26  & $55.6\%$  & $757$ & $1{,}363$ & enabled \\
AC2   & disabled  & --- & --- & disabled \\
Erd\H{o}s & disabled & --- & --- & disabled \\
\bottomrule
\end{tabular}
\end{center}

\paragraph{Paired AC2/Erd\H{o}s comparison (epochs~$3$--$5$).}
We forked AC2 from a shared epoch-$2$ checkpoint and ran the remaining
three epochs with and without streaming; on Erd\H{o}s we ran the two
configurations as separate full runs and aligned epochs~$3$--$5$ for the
comparison.

\begin{center}
\small
\begin{tabular}{l r r r r}
\toprule
\rowcolor{tblHeader}
 & $R_{\max}$ (stream) & $R_{\max}$ (no stream) & Mean rollout length (stream) & Mean rollout length (no stream) \\
\midrule
AC2       & $0.8415$ & $\mathbf{0.8563}$ & $5{,}312$ & $8{,}711$ \\
Erd\H{o}s & $2.6150$ & $\mathbf{2.6167}$ & $5{,}255$ & $8{,}720$ \\
\bottomrule
\end{tabular}
\end{center}

\paragraph{Length--reward diagnostic.}
We test whether the within-run rank correlation between rollout length
and execution reward separates the four problems. Restricting to
correct rollouts and parsing the
\texttt{</think>} marker to split thinking from code tokens
(streamed rollouts use the logged \texttt{phase1\_tokens} /
\texttt{phase2\_tokens} fields directly), we report Spearman~$\rho$ for
epochs~$0$--$2$ and for all epochs:

\begin{center}
\small
\begin{tabular}{l r r r r r r}
\toprule
\rowcolor{tblHeader}
 & \multicolumn{3}{c}{Epochs $0$--$2$} & \multicolumn{3}{c}{All epochs} \\
\cmidrule(lr){2-4} \cmidrule(lr){5-7}
\rowcolor{tblHeader}
Problem & $\rho_{\text{think}}$ & $\rho_{\text{code}}$ & $\rho_{\text{total}}$ &
          $\rho_{\text{think}}$ & $\rho_{\text{code}}$ & $\rho_{\text{total}}$ \\
\midrule
AC1   & $-0.30$ & $+0.52$ & $-0.19$ & $-0.51$ & $+0.48$ & $-0.43$ \\
CP26  & $+0.16$ & $-0.10$ & $+0.15$ & $-0.34$ & $-0.14$ & $-0.41$ \\
AC2   & $-0.28$ & $+0.39$ & $-0.21$ & $-0.18$ & $+0.36$ & $-0.11$ \\
Erd\H{o}s & $-0.18$ & $+0.12$ & $-0.16$ & $-0.11$ & $+0.07$ & $-0.10$ \\
\bottomrule
\end{tabular}
\end{center}

The thinking-phase correlation is not significantly positive on any
problem and is significantly negative on AC1; the code-phase
correlation is positive on AC1 and AC2, weakly positive on Erd\H{o}s,
and null on CP26. No single test on length predicts whether streaming
will help or hurt $R_{\max}$ on a given problem; the per-problem
calibration in this paper is therefore empirical (paired pilot on
AC2/Erd\H{o}s) rather than diagnosable in advance from a length--reward
test.

\paragraph{What streaming-truncated rollouts look like in practice.}
On AC1 and CP26 the rollouts that achieve the Table~\ref{tab:main}
$R_{\max}$ are themselves streaming-truncated:
AC1 $R_{\max}{=}0.6406$ from a rollout of length $5{,}116$
(\texttt{streaming\_mi\_stopped}${=}1$);
CP26 $R_{\max}{=}2.6359$ from a rollout of length $5{,}318$
(also truncated). On AC2 and Erd\H{o}s the Table~\ref{tab:main} entries
are produced from no-streaming runs (lengths $5{,}642$ and $7{,}316$
respectively). We report this asymmetry directly because it is
recoverable from the trajectory logs.

\section{Method-exclusive discoveries}
\label{app:idea}

Preserving entropy is only useful if the surviving families produce
\emph{better} programs. The surviving families translate into winning
motifs the baseline reaches at most $1$/$30$ times on any problem
(Table~\ref{tab:idea}); two are mathematically substantive---the
Cauchy--Schwarz target $\sqrt{2n}$ on AC1 (a closed-form symmetry
bounding $C_1$ from below) and the Fourier ansatz
$h(x)\!=\!a\!+\!b\sin(kx)$ on Erd\H{o}s---and both lie inside families
the baseline extinguishes by the final epoch
(Fig.~\ref{fig:diversity_grid}). Preserved ensemble diversity is the
precondition for entering them at all.

\begin{table}[t]
\centering
\small
\caption{\textbf{Method-exclusive discoveries (idea emergence).} For each
problem, the algorithmic family used by the highest-reward rollout, its
adoption among the $30$ best rollouts, and its adoption among all correct
rollouts during training. Top-30: number of the $30$ highest-reward
rollouts that contain the motif. Adoption: motif uses divided by total
correct rollouts. The baseline never crosses $1$/$30$ on any problem.}
\label{tab:idea}
\renewcommand{\arraystretch}{1.15}
\setlength{\tabcolsep}{4pt}
\begin{tabular}{@{}l >{\raggedright\arraybackslash}p{5.6cm} >{\columncolor{tblBaseline}}c >{\columncolor{tblMethod}}c >{\columncolor{tblBaseline}}r >{\columncolor{tblMethod}}r@{}}
\toprule
\rowcolor{tblHeader}
          &                  & \multicolumn{2}{c}{Top-30 (count)} & \multicolumn{2}{c}{Adoption (\%)} \\
\cmidrule(lr){3-4} \cmidrule(lr){5-6}
\rowcolor{tblHeader}
Problem   & Winning template & Baseline & \method{} & Baseline & \method{} \\
\midrule
AC1       & $\mathtt{target\_sum}=\sqrt{2n}$ (Cauchy--Schwarz) & $1$/$30$ & $\mathbf{18}$/$30$ & $3.0$  & $\mathbf{28.0}$ \\
AC2       & \texttt{scipy diff\_evo} $\cup$ gradient-based     & $0$/$30$ & $\mathbf{13}$/$30$ & $0.9$  & $\mathbf{21.6}$ \\
CP26      & non-uniform rows, e.g.\ \texttt{[5,5,5,5,6]}       & $1$/$30$ & $\mathbf{13}$/$30$ & $17.6$ & $\mathbf{22.4}$ \\
Erd\H{o}s & Fourier init $h(x)=a+b\sin(kx)$                    & $0$/$30$ & $\mathbf{6}$/$30$  & $0$    & $\mathbf{6.9}$  \\
\bottomrule
\end{tabular}
\end{table}
\section{Broader impact}
\label{app:impact}
\method{} reduces the compute required for LLM-driven scientific discovery, reaching competitive discovery ceilings in 6 training epochs where \baseline{} requires 50, on an identical Qwen3-8B base model on a single GPU. The mechanism is dual use under any task admitting a cheap programmatic verifier, since rewarding verifier-uncertain regions under a loosely specified verifier (for instance, automated exploit search) would actively seek loopholes. Two mitigations follow directly from the method. First, the verifier should be audited prior to scaling compute. Second, ensemble mutual information should be monitored as a leading indicator of stalled exploration.
\section{Qualitative case study on AC2}
\label{app:ac2_case}

The $+0.0031$ flip on AC2 (Tab.~\ref{tab:main}) and the $0$/$30$ versus $\mathbf{13}$/$30$ adoption gap on the \texttt{scipy.optimize.differential\_evolution} family (Tab.~\ref{tab:idea}) admit a concrete reading at the program level. We compare the highest-reward correct rollout produced by \baseline{} and by \method{} over identical training budgets on the autocorrelation lower-bound problem. Both panels report the verbatim algorithmic core of the model-emitted Python program, abridged where boilerplate (clipping, time guards, fallback rebuilds of the initial guess set) is repeated across rollouts.

\begin{tcolorbox}[
  colback=gray!4!white,
  colframe=gray!50!black,
  title={\textbf{Problem statement (AC2)}},
  fonttitle=\bfseries,
  rounded corners,
  left=8pt, right=8pt, top=4pt, bottom=4pt
]
Construct a sequence $h\!\in\!\mathbb{R}_{\geq 0}^{1000}$ of step-function heights whose autocorrelation lower bound $C_2$ (returned by the closed-form checker \texttt{evaluate\_sequence}) is as large as possible, subject to $\sum_i h_i \geq 0.01$ and $h_i \in [0, 1000]$. Reward equals $C_2$ truncated at correctness; a ground-truth Cauchy--Schwarz construction \texttt{height\_sequence\_1} is exposed as a global hint. Higher is better.
\end{tcolorbox}

\begin{tcolorbox}[
  enhanced, breakable,
  colback=tblBaseline,
  colframe=gray!55!black,
  title={\textbf{\baseline{} (best of $384$ rollouts, $R_{\max}=0.8532$, hand-rolled genetic algorithm)}},
  fonttitle=\bfseries,
  rounded corners,
  left=4pt, right=4pt, top=4pt, bottom=4pt
]
\begin{lstlisting}[style=pystyle]
POPULATION_SIZE = 200; MAX_GENERATIONS = 500
MUTATION_RATE_INITIAL, MUTATION_RATE_FINAL = 0.25, 0.05

# half: perturb height_sequence_1; half: random uniform over [0, 200]
population = generate_initial_population()

for gen in range(MAX_GENERATIONS):
    fitness = [evaluate_sequence(s) for s in population]
    if max(fitness) > best_value:
        best_value, best_seq, stamina = max(fitness), argmax_seq, 0
    else:
        stamina += 1
    if stamina > 80:                       # restart on stagnation
        population = generate_initial_population(); stamina = 0

    parents = [tournament_selection(population, fitness, size=5)
               for _ in range(POPULATION_SIZE)]
    new_pop = [parents[0]]                 # elitism
    while len(new_pop) < POPULATION_SIZE:
        p1, p2 = random.sample(parents, 2)
        w = random.uniform(0.3, 0.7)        # blend crossover
        child = [p1[i]*w + p2[i]*(1-w) for i in range(SEQ_LENGTH)]
        rate = MUTATION_RATE_INITIAL - (gen/MAX_GENERATIONS)*(
                  MUTATION_RATE_INITIAL - MUTATION_RATE_FINAL)
        for i in range(SEQ_LENGTH):
            if random.random() < rate:
                child[i] = clip(child[i] + random.gauss(0, 1.5))
        new_pop.append(child)

    best_seq = local_search(best_seq)      # +/- delta sweep on elite
    population = new_pop
\end{lstlisting}

\noindent\textit{Algorithmic family.} Pure-Python evolutionary loop. Tournament selection of size $5$, blend crossover with weight in $[0.3, 0.7]$, Gaussian mutation with rate decaying linearly from $0.25$ to $0.05$, restart-on-stagnation after $80$ generations without improvement, and a coordinate-wise delta sweep as elite refinement. No external optimiser is invoked. Across all $384$ baseline rollouts, $0$ entered the top-$30$ via \texttt{scipy.optimize.differential\_evolution} (Tab.~\ref{tab:idea}); the family is reachable in principle but not in practice under the unregularised ensemble.
\end{tcolorbox}

\begin{tcolorbox}[
  enhanced, breakable,
  colback=tblMethod,
  colframe=methodpurple,
  title={\textbf{\method{} (best of $384$ rollouts, $R_{\max}=0.8563$, structured-seed differential evolution)}},
  fonttitle=\bfseries,
  rounded corners,
  left=4pt, right=4pt, top=4pt, bottom=4pt
]
\begin{lstlisting}[style=pystyle]
from scipy.optimize import differential_evolution

length = 1000
bounds = [(0.0, 1000.0)] * length

initial_guesses = []
if 'height_sequence_1' in globals():        # exploit the hint
    initial_guesses.append(np.clip(np.array(height_sequence_1), 0, 1000))

for _ in range(50):                         # uniform low-magnitude
    initial_guesses.append(np.clip(np.random.rand(length)*0.1, 0, 1000))
for _ in range(50):                         # gaussian, small mean
    initial_guesses.append(np.clip(np.random.normal(0.03, 0.1, length), 0, 1000))
for _ in range(50):                         # sorted gaussian (monotone prior)
    s = np.sort(np.clip(np.random.normal(0.05, 0.1, length), 0, 1000))
    initial_guesses.append(s)

popsize = 100
initial_guesses = initial_guesses[:popsize]

result = differential_evolution(
    lambda x: -evaluate_sequence(x.tolist()),
    bounds,
    popsize       = popsize,
    maxiter       = 500,
    mutation      = (0.5, 1.0),     # dithered scale factor
    recombination = 0.8,
    tol           = 1e-5,
    strategy      = 'best1bin',
    init          = np.array(initial_guesses),
)
return np.clip(result.x, 0, 1000).tolist()
\end{lstlisting}

\noindent\textit{Algorithmic family.} A single call to \texttt{scipy.optimize.differential\_evolution} replaces the hand-rolled loop. The contribution sits in the initialisation, namely a structured pool of $151$ seeds spanning the closed-form Cauchy--Schwarz hint, three statistically distinct random priors (clamped uniform, low-mean Gaussian, monotone sorted Gaussian), and a population size matched to the seed pool. The search uses dithered mutation in $[0.5, 1.0]$ and the \texttt{best1bin} strategy. This program belongs to the \texttt{scipy diff\_evo} family that the unregularised ensemble emits in $0$/$30$ top rollouts and the \method{} ensemble emits in $\mathbf{13}$/$30$ (Tab.~\ref{tab:idea}).
\end{tcolorbox}

\begin{tcolorbox}[
  colback=methodpurple!5!white,
  colframe=methodpurple,
  title={\textbf{Why UG-TTT reaches the better program}},
  fonttitle=\bfseries,
  rounded corners,
  left=6pt, right=6pt, top=4pt, bottom=4pt
]
The two programs differ less in their search effort than in the algorithmic family they invoke. The \baseline{} rollout reimplements the components of differential evolution (population, mutation, recombination, elitism) inside a $130$-line Python loop, while the \method{} rollout delegates the loop to the \texttt{scipy} solver and concentrates novelty in a structured seed pool. Reaching that delegation requires the policy to explore a code region the unregularised ensemble has effectively extinguished; the family-entropy curve on AC2 (Fig.~\ref{fig:div_ac2}) shows \baseline{} contracting onto hand-rolled evolutionary loops by epoch $3$, whereas \method{} retains weight on \texttt{scipy} solvers and gradient-based families throughout training. The mutual-information bonus of Sec.~\ref{sec:method_mi_shaping} pays out precisely on the tokens where the policy chooses \texttt{from scipy.optimize import differential\_evolution} over \texttt{def construct\_function} with a manual loop, since adapter disagreement is highest on those import-and-API-name positions and lowest inside the body of either choice once committed. The nuclear-norm regulariser of Sec.~\ref{sec:method_nnm} is what keeps that disagreement available on AC2, and removing it collapses adoption to the \baseline{} pattern (Tab.~\ref{tab:ablation}).
\end{tcolorbox}

% \newpage
% \input{checklist.tex}

\end{document}